\documentclass{article}
\usepackage[T1]{fontenc}
\usepackage{valenteam}
\usepackage{microtype}
\usepackage{amsmath,amsthm}
\usepackage{hyperref}
\usepackage{xspace}

\usepackage{textcomp}
\makeatletter
\DeclareFontFamily{OMS}{zi4}{}
\DeclareFontShape{OMS}{zi4}{m}{n}{<->ssub*cmsy/m/n}{}
\DeclareFontShape{OMS}{zi4}{b}{n}{<->ssub*cmsy/b/n}{}
\DeclareFontShape{OMS}{zi4}{bx}{n}{<->ssub*cmsy/b/n}{}
\makeatother

\usepackage{amsmath, amssymb, mathtools, nicefrac}
\usepackage{booktabs, multirow, array, makecell, adjustbox, colortbl, tabularx}
\usepackage{graphicx, subcaption, float, wrapfig, placeins}
\usepackage{pifont}
\usepackage{caption}
\usepackage{listings}
\usepackage[capitalize]{cleveref}

\definecolor{vtgreen}{HTML}{1B9E5A}

\newcommand{\up}[1]{\textsubscript{\textcolor{vtgreen}{+#1}}}

\lstdefinestyle{vtcode}{
    basicstyle=\ttfamily\fontsize{8pt}{9.5pt}\selectfont,
    keywordstyle=\color{vtdeepblue}\bfseries,
    commentstyle=\color{vtgray},
    stringstyle=\color{vtgreen},
    backgroundcolor=\color{vtpaleblue},
    frame=none,
    breaklines=true,
    breakatwhitespace=true,
    breakindent=0pt,
    columns=fullflexible,
    keepspaces=true,
    showstringspaces=false,
    escapeinside={(*@}{@*)},
    numbers=none,
    xleftmargin=5pt,
    xrightmargin=5pt,
}
\newtcolorbox{VTbox}[2][]{
    enhanced, breakable,
    colback=white,
    colframe=vtblue,
    colbacktitle=vtblue,
    coltitle=white,
    fonttitle=\bfseries\small,
    boxrule=0.8pt,
    arc=2pt,
    left=6pt, right=6pt, top=5pt, bottom=5pt,
    title=#2, #1
}
\newtcolorbox{VTnote}[1][]{
    enhanced, breakable,
    colback=vtpaleblue,
    colframe=vtpaleblue,
    borderline west={2.5pt}{0pt}{vtblue},
    boxrule=0pt,
    arc=0pt,
    left=8pt, right=6pt, top=4pt, bottom=4pt,
    #1
}

\usepackage{tikz}
\usetikzlibrary{arrows.meta,positioning,fit,calc,backgrounds,shapes.misc,decorations.pathreplacing}

\theoremstyle{definition}
\newtheorem{definition}{Definition}
\usepackage{relsize}
\DeclareFontSeriesDefault[sf]{bf}{bx}
\usepackage{algorithm}
\usepackage{algorithmic}

\usepackage{etoolbox}
\AtBeginEnvironment{table}{\renewcommand{\arraystretch}{1.06}}

\let\up\relax

\newcommand{\roller}{Roller}

\makeatletter
\newcommand{\reportcodesize}{\ifdim\f@size pt<8.4pt\relax\else\relsize{-1}\fi}
\makeatother
\newcommand{\code}[1]{{\reportcodesize\texttt{#1}}}
\newcommand{\file}[1]{{\reportcodesize\texttt{#1}}}   
\newcommand{\flag}[1]{{\reportcodesize\texttt{-{}-#1}}} 

\newcommand{\na}{\textcolor{black!35}{n/a}}

\newcommand{\up}{$\uparrow$}

\newcommand{\yes}{\textcolor{raqua}{\ding{51}}}
\newcommand{\no}{\textcolor{black!35}{\ding{55}}}

\newcommand{\tablehead}{\rowcolor{rviolet!7}}
\newcommand{\tablesource}[1]{\textsuperscript{\citeyearpar{#1}}}
\newcommand{\resultbar}[4][rviolet]{%
  \begingroup
  \pgfmathsetlengthmacro{\resultwidth}{#2 / #3 * 3.9em}%
  \makebox[3.9em][l]{%
    \rlap{\textcolor{#1!7}{\rule[-.25ex]{3.9em}{1.8ex}}}%
    \rlap{\textcolor{#1!24}{\rule[-.25ex]{\resultwidth}{1.8ex}}}%
    \makebox[3.9em][r]{#4\hspace{.25em}}%
  }%
  \endgroup
}

\newcommand{\Tset}{\mathcal{T}}
\newcommand{\Mset}{\mathcal{M}}

\DeclareMathOperator{\digest}{digest}

\definecolor{rblue}{HTML}{2A78D6}
\definecolor{rorange}{HTML}{EB6834}
\definecolor{raqua}{HTML}{1BAF7A}
\definecolor{rgreen}{HTML}{00866F}     
\definecolor{rviolet}{HTML}{8B5CC9}
\definecolor{rink}{HTML}{52514E}
\definecolor{rmuted}{HTML}{8A8984}
\definecolor{rgrid}{HTML}{E4E3DF}
\definecolor{rhost}{HTML}{F4F2FC}      
\definecolor{rbox}{HTML}{EEF8F4}       
\definecolor{rlight}{HTML}{EEF3FB}

\tikzset{
  rbox/.style={draw=rmuted!70, rounded corners=2pt, fill=white, align=center, inner sep=3pt,
               font=\scriptsize, line width=0.4pt},
  rdash/.style={rbox, dashed},
  rarr/.style={-{Stealth[length=3.2pt]}, draw=rink, line width=0.5pt},
  rcall/.style={-{Stealth[length=3.2pt]}, draw=rgreen, line width=0.7pt},
  rdarr/.style={rarr, dashed},
  rlbl/.style={font=\tiny, text=rink, inner sep=1pt},
  rgroup/.style={rounded corners=4pt, line width=0.4pt, inner sep=5pt},
}

\newtcolorbox{casebox}[2]{enhanced, frame hidden, colback=#2!4, arc=3pt,
  left=6pt, right=6pt, top=3pt, bottom=4pt, fontupper=\fontsize{8.1}{9.6}\selectfont,
  before upper={\raggedright}, colbacktitle=#2!15, coltitle=rink, fonttitle=\footnotesize\bfseries,
  toptitle=2pt, bottomtitle=2pt, titlerule=0pt, title={#1}}
\newcommand{\turn}[1]{\textsf{\textbf{#1}}\enspace}          
\newcommand{\tool}[1]{\texttt{\fontsize{6.9}{8}\selectfont #1}}

\renewcommand{\VTlogo}[1][1.05em]{%
    \includegraphics[height=#1]{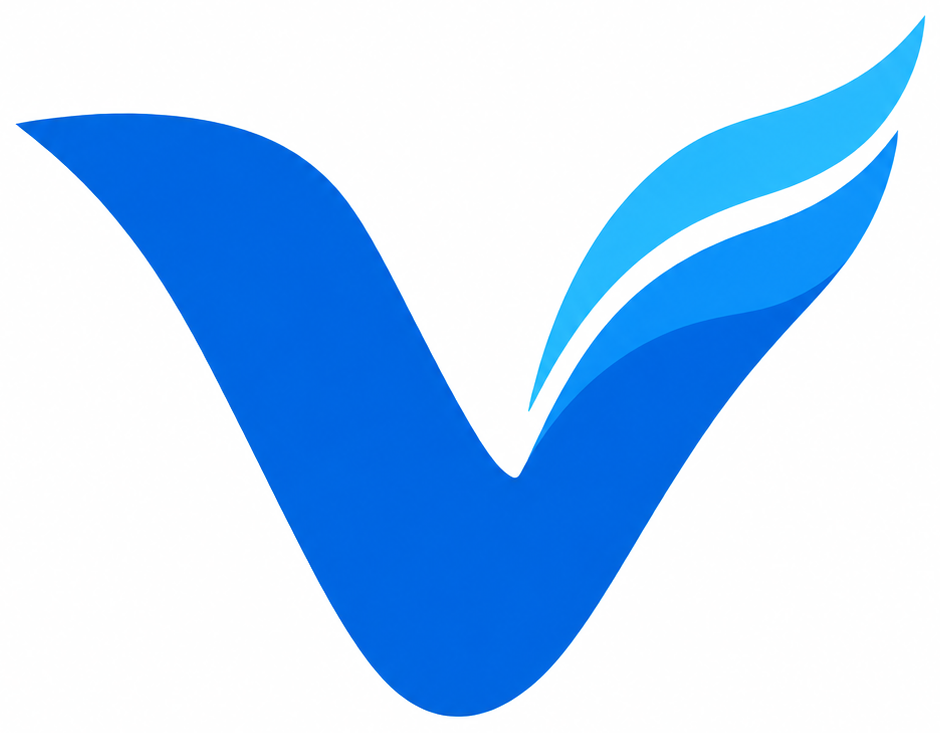}%
}

\date{November 2026}
\runningtitle{Roller: Agent Trajectory Collection Across Models and Harnesses}
\ghlink{https://github.com/Valen-Team/Roller}
\title{Roller: Reproducible Agent Trajectory Collection Across Models and Harnesses}
\affiliations{Shanghai Jiao Tong University, Fudan University,\\Sun Yat-sen University}
\hypersetup{
    colorlinks=true, citecolor=vtdeepblue, linkcolor=vtdeepblue, urlcolor=vtdeepblue,
    pdftitle={Roller: Reproducible Agent Trajectory Collection Across Models and Harnesses},
    pdfauthor={Valen Team},
    pdfkeywords={agents, trajectories, agent harness, supervised fine-tuning, evaluation, reproducibility}
}

\begin{document}
\valenfrontmatter{%
    \begin{abstract}
Agentic development tools such as Claude Code, Codex and Google Antigravity use runtimes that manage tools and
context. A runtime's session log can therefore differ from the requests sent to the model.
This difference matters when trajectories are reused for training and evaluation. We present Roller, an
open-source system for collecting trajectories across eight supported runtimes. Roller binds each rollout to its
inputs, configuration and execution environment. It links native sessions to normalized steps and optional
model-call evidence. For compacted rollouts with sufficient capture, it reconstructs training windows and
marks which assistant messages contribute to the loss. In a reasoning audit spanning 196 configurations,
Roller identified eight explicit conflicts and excluded all eight from the training aggregate. On held-out
slide generation, blanking teacher reasoning above 2,000 characters before fine-tuning raised the grader score
from 9.40 to 9.62 with 20 of 20 tasks reported complete. Blanking all reasoning yielded 9.70 with seven
reported completions. These experiments illustrate why task scores, execution status and training eligibility
require separate measurements.
\end{abstract}
}
\section{Introduction}
\label{sec:intro}

Language-model agents act through runtimes that assemble prompts, expose tools and manage context. We call
these runtimes \emph{Harnesses}. The choice of Harness can substantially affect the measured performance of
a fixed model. On Terminal-Bench 2.0, GPT-5 resolves 33.9\% of tasks with Mini-SWE-Agent and 49.6\% with Codex
CLI \citep{terminalbench}. WildClawBench reports differences of up to 18 percentage points across Harnesses
\citep{wildclawbench}. For trajectory collection, the same dependence means that training and analysis benefit
from records of the Harness, tools and context management used during execution
\citep{luo2025agentlightning,he2026agentlightningv1}.

\begin{figure}[tbp]
\centering
\begin{tikzpicture}[
  x=0.82cm, y=0.51cm,
  row label/.style={anchor=west, inner sep=0pt, font=\small, text=rink},
  group label/.style={row label, font=\small\bfseries},
  note/.style={inner sep=1pt, font=\scriptsize, text=rink},
  message/.style={rounded corners=1.5pt, minimum width=0.65cm, minimum height=0.40cm,
    inner sep=0pt, font=\footnotesize, text=rink, fill=#1},
  system/.style={message=black!10},
  user/.style={message=rblue!16},
  assistant/.style={message=raqua!16},
  observation/.style={message=rorange!17},
  summary/.style={message=rviolet!20},
  target/.style={message=rgreen!85!black, text=white, font=\footnotesize\bfseries},
  wrong/.style={target, fill=red!62!black}
]
  \foreach \y in {-0.90,-4.24,-5.89}
    \draw[rgrid, line width=0.45pt] (0,\y) -- (14.30,\y);

  \node[group label] at (0,0) {Harness session};
  \foreach \x/\tone/\itemtext in {
    4.75/user/U,5.75/assistant/A_1,6.75/observation/O_1,7.75/assistant/A_2,
    8.75/observation/O_2,9.75/summary/\Sigma,10.75/assistant/A_3,
    11.75/observation/O_3,12.75/assistant/A_4}
    \node[\tone] at (\x,0) {$\itemtext$};
  \node[note, text=rviolet!75!black] at (9.75,0.68) {Compaction};

  \node[group label] at (0,-1.50) {Captured model calls};
  \node[row label] at (0,-2.50) {Before compaction};
  \node[row label] at (0,-3.45) {After compaction};
  \foreach \y/\secondtone/\secondmessage/\prioranswer/\toolresult/\answer in {
    -2.50/user/U/A_1/O_1/A_2,-3.45/summary/\Sigma/A_2/O_2/A_3} {
    \node[system] at (4.75,\y) {$S$};
    \node[\secondtone] at (5.75,\y) {$\secondmessage$};
    \node[assistant] at (6.75,\y) {$\prioranswer$};
    \node[observation] at (7.75,\y) {$\toolresult$};
    \draw[rarr] (8.22,\y) -- (8.83,\y);
    \node[assistant] at (9.35,\y) {$\answer$};
  }
  \node[note, anchor=west] at (9.90,-2.50) {Model response};

  \node[group label] at (0,-5.07) {Session-only SFT};
  \foreach \x/\tone/\itemtext in {4.75/system/S,5.75/user/U,6.75/assistant/A_1,
    7.75/observation/O_1,8.75/assistant/A_2,9.75/observation/O_2,10.75/wrong/A_3}
    \node[\tone] at (\x,-5.07) {$\itemtext$};
  \node[note, anchor=west, text=red!62!black] at (11.32,-5.07) {Context mismatch};

  \node[group label] at (0,-6.46) {Roller SFT export};
  \node[row label] at (0,-7.54) {Segment 1};
  \node[row label] at (0,-8.73) {Segment 2};
  \foreach \x/\tone/\itemtext in {
    4.75/system/S,5.75/user/U,6.75/target/A_1,7.75/observation/O_1,8.75/target/A_2}
    \node[\tone] at (\x,-7.54) {$\itemtext$};
  \foreach \x/\tone/\itemtext in {
    4.75/system/S,5.75/summary/\Sigma,6.75/assistant/A_2,7.75/observation/O_2,
    8.75/target/A_3,9.75/observation/O_3,10.75/target/A_4}
    \node[\tone] at (\x,-8.73) {$\itemtext$};
  \foreach \x/\y/\index in {6.25/-7.54/2,8.25/-8.73/4} {
    \draw[rink, densely dashed, line width=0.55pt] (\x,\y-0.40) -- (\x,\y+0.43);
    \node[note, anchor=west] at (11.32,\y) {$t=\index$};
  }
  \node[target] at (4.75,-9.83) {$A$};
  \node[note, anchor=west] at (5.28,-9.83) {Included in loss};
  \node[assistant] at (9.15,-9.83) {$A$};
  \node[note, anchor=west] at (9.68,-9.83) {Outside the loss};
\end{tikzpicture}
\caption{An illustrative rollout with context compaction. The Harness replaces early history with a summary
$\Sigma$ and keeps the recent exchange $(A_2,O_2)$. Replaying the session as a single SFT record would train
$A_3$ on pre-compaction history. \roller{} starts each segment with the actual prompt from its first captured
call. The dashed line marks the zero-based \code{target\_start\_index} ($t$), the first position eligible for assistant
supervision. The replayed $A_2$ conditions $A_3$ in segment 2 and contributes to the loss only in segment 1.
$S$ denotes the system prompt, $U$ the user message, $A$ assistant responses, and $O$ tool observations.}
\label{fig:teaser}
\end{figure}
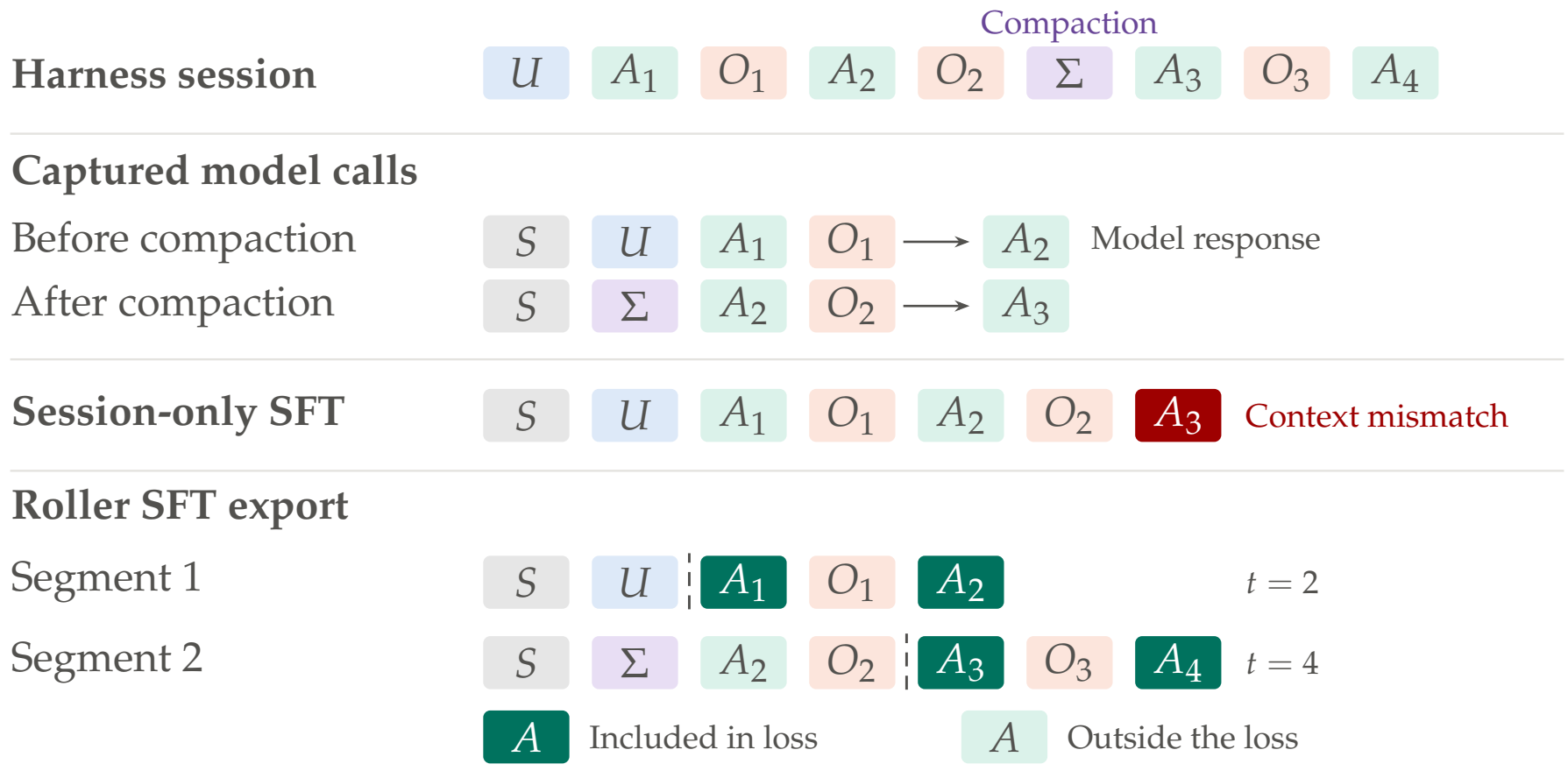

Using these trajectories for training or evaluation requires evidence about both the model interaction and its
collection conditions. A native session records events for the runtime. The model request can contain a
different history after compaction, tool-result pruning or system prompt injection
\citep{lu2025supo,ye2025agentfold,kang2025acon}. \Cref{fig:teaser} illustrates the consequence for supervision.
Replaying the session as a linear example can pair a response with a prompt that was never sent. Configuration
and outcome records have a similar gap. In one rollout, the task passed even though every outgoing request
carried \code{high} reasoning effort after the run requested \code{low} (\cref{fig:cases}, panel (a)).
Interrupted runs introduce another source of variation when task inputs, runtime images or model routing
change before collection resumes.

Evaluation frameworks provide isolated execution and benchmark grading
\citep{harbor,shi2026harboradapters,inspectai}, and trajectory schemas standardize records for reuse
\citep{song2025adp}. We address the collection problem of linking a Harness session and its outputs to the
model interactions and configuration that produced them. We present \roller{}, an open-source collection
system that runs tasks across eight Harnesses and retains native sessions, normalized steps and optional
model-call evidence \citep{roller2026repo}.

This work makes three contributions.

\begin{itemize}
\item \textbf{Reproducible collection across Harnesses.} Roller binds each rollout to snapshotted inputs and a
run lock, and checks those conditions on resume. Benchmark graders use the live workspace or files collected
before teardown (\cref{sec:system}).
\item \textbf{Training records grounded in captured evidence.} Roller reconstructs training windows from
captured requests, marks the positions eligible for supervision, and uses native identifiers to attribute
sub-agent calls. Audits use recorded evidence to determine which samples enter the training aggregate
(\cref{sec:trajectories}).
\item \textbf{Empirical analysis of trajectory quality.} In a 196-cell reasoning audit, Roller identified
eight explicit conflicts and excluded all eight from the SFT aggregate. Benchmark and fine-tuning results
illustrate how task scores and completion can diverge. On held-out slides,
blanking reasoning above 2,000 characters raised the grader score from 9.40 to 9.62 with all 20 tasks
reported complete. Blanking all reasoning yielded 9.70 with seven reported completions
(\cref{sec:experiments}).
\end{itemize}

\section{Related Work}
\label{sec:related}

\paragraph{Agent runtimes.}
ReAct \citep{yao2023react} and Toolformer \citep{schick2023toolformer} established the loop of reasoning, tool
calls and observations that agent runtimes now implement. SWE-agent \citep{yang2024sweagent} showed that the
interface a runtime gives the model changes what the model can do, and OpenHands \citep{wang2025openhands} packages
such interfaces as an open platform with sandboxed execution. AutoGen \citep{wu2024autogen} and MetaGPT
\citep{hong2024metagpt} coordinate several agents, which is now common in commercial runtimes as sub-agent
delegation. Production Harnesses such as Claude Code \citep{claudecode}, Codex \citep{codexcli}, OpenClaw
\citep{openclaw}, Hermes Agent \citep{hermesagent}, OpenCode \citep{opencode}, the DeepSeek Harness \citep{deepseekharness} and Pi \citep{piagent} add context
compaction, skills and persistent memory on top of this loop. \roller{} collects trajectories from these
Harnesses at pinned versions while preserving their native tools and context-management behavior.
Google Antigravity extends agentic coding workflows to an integrated development platform
\citep{antigravity}.

\paragraph{Benchmarks and evaluation harnesses.}
SWE-bench \citep{jimenez2024swebench}, AgentBench \citep{liu2024agentbench}, WebArena \citep{zhou2024webarena},
OSWorld \citep{xie2024osworld} and $\tau$-bench \citep{yao2024taubench} define environments and graders for
agents, and MINT \citep{wang2024mint} and $\tau$-bench use a language model as a simulated user. Inspect
\citep{inspectai} and Harbor \citep{harbor}, the framework behind Terminal-Bench \citep{terminalbench}, run agents in
containers and score their final state, and Harbor's adapters connect its agents to more than 80 benchmarks
\citep{shi2026harboradapters}. Roller uses benchmark grading alongside trajectory collection and supports
Harbor's ATIF format for interchange. Work on evaluation methodology
\citep{liang2023helm,biderman2024lessons,kapoor2024agents} has shown how small differences in prompts, versions and
cost accounting change reported agent scores. The run lock and per-sample provenance in \roller{} respond to the
same concern.

\paragraph{Trajectories as training data.}
FireAct \citep{chen2023fireact}, AgentTuning \citep{zeng2024agenttuning}, Agent-FLAN \citep{chen2024agentflan} and
ToolLLM \citep{qin2024toolllm} fine-tune models on agent trajectories, and SWE-Gym \citep{pan2025swegym} and
SWE-smith \citep{yang2025swesmith} scale this to software engineering. Kimi K2 \citep{kimi2025k2} and GLM-4.5
\citep{glm2025glm45} report large agentic data pipelines for pre- and post-training. The Agent Data Protocol
\citep{song2025adp} converts many existing trajectory datasets into one schema after the fact. Agent Lightning
\citep{luo2025agentlightning,he2026agentlightningv1} and the frameworks that adopted its design connect a deploy-time
Harness to RL training through a model-endpoint proxy, so the trainer sees only request--response pairs; \roller{}'s
request capture records the same pairs and keeps them as a dataset. Methods that train agents to summarize or compress
their own context \citep{lu2025supo,ye2025agentfold,kang2025acon} build training samples from the managed context,
while Roller reconstructs training segments from the requests that existing Harnesses send. This gives
the collection pipeline evidence about the prompt used for each supervised response.

\paragraph{Observability and provenance.}
LLM tracing tools such as LangSmith \citep{langsmith}, Langfuse \citep{langfuse} and the OpenTelemetry generative-AI
conventions \citep{otelgenai} record spans and model calls from inside an application that the developer
instruments. \roller{} gathers evidence from a Harness's native logs and events and, when available,
from the model-call boundary. For inputs, \roller{} uses content addressing
from reproducible build systems \citep{merkle1987digital,dolstra2004nix}. For outputs, it follows the view of
dataset documentation work \citep{gebru2021datasheets,longpre2023provenance} that provenance belongs to the record
itself.

\section{Trajectories, Requests and Training Records}
\label{sec:problem}

\paragraph{Terms.}
A \emph{task} $\tau$ holds a prompt or a sequence of user turns, input files, Context files (such as
\file{AGENTS.md}, \file{MEMORY.md} or a skills directory), setup commands and an output directory. A
\emph{Harness} $h$ is an agent runtime with its own tools, context management and model protocol. A
\emph{rollout} is one attempt to run task $\tau$ with model $m$ on Harness $h$; a \emph{run} is a named batch of
rollouts over $\Tset \times \Mset$, and a \emph{sample} is the directory a rollout leaves behind.

\paragraph{Three records of one rollout.}
During a rollout the Harness issues model calls $1, \dots, K$. Call $k$ sends a request context $x_k$ and receives a
response $y_k$. Between calls the Harness executes tools and receives observations $o_k$, and in a multi-turn task
it receives new user messages $u_k$. These pairs $(x_k, y_k)$ are the facts a training set should reproduce. The
Harness writes a different record, its session $S$: a list of events in a Harness-specific schema. A training
record is a third object, a message list $L = (\ell_1, \dots, \ell_n)$ in which selected assistant messages are
targets and the remaining messages are context. A linear projection of $S$ is faithful only if it preserves
the context for each supervised response.

\begin{definition}[Append-only window]
\label{def:window}
Calls $k, k{+}1, \dots, l$ form an append-only window when, for every $i$ with $k \le i < l$, the request $x_{i+1}$
equals $x_i$ followed by $y_i$, then $o_i$ and $u_i$, up to representation in the training format.
\end{definition}

Within a window, one record can start with $x_k$ and supervise $y_k, \dots, y_l$. Each target keeps the same
conditioning context as in its per-call example, and the record stores shared prefixes once. Compaction,
tool-result pruning, system prompt rewrites and dropped reasoning blocks can break the append-only condition.
A linear projection of the session can then condition later targets on a history that was never sent
(\cref{fig:teaser}).
When a context reset has sufficient captured evidence, \roller{} partitions the main agent's calls into windows.
Each record starts from the first captured request in its window, represented in the training format
(\cref{sec:segments}). Short rollouts with append-only history use the ordinary linear export.

The session supplies native tool events, compaction markers and sub-agent identities where the Harness records
them. Roller's capture layer links that evidence to individual model calls. The recorded wire body establishes
the request sent to the service. Tokenization, truncation and routing inside the service require separate
evidence. \Cref{tab:layers} shows which artifacts preserve native events, model traffic and normalized steps.
Roller uses those artifacts to build the training records described in \cref{sec:trajectories}.

\section{Roller Architecture}
\label{sec:system}

\begin{figure}[tbp]
\centering
\begin{tikzpicture}[
  x=1cm, y=1cm,
  card/.style={fill=#1, rounded corners=2.5pt, minimum height=0.74cm,
    inner xsep=2pt, inner ysep=2pt, align=center,
    font=\fontsize{7.7}{8.6}\selectfont\bfseries, text=rink},
  optional/.style={card=white, dashed, line width=0.6pt},
  boundary/.style={rounded corners=5pt, line width=0.55pt},
  label text/.style={font=\fontsize{6.6}{7.4}\selectfont, text=rink, inner sep=1.4pt},
  flow/.style={-{Stealth[length=3.4pt,width=3pt]}, draw=rink!75, line width=0.55pt},
  exchange/.style={{Stealth[length=3.4pt,width=3pt]}-{Stealth[length=3.4pt,width=3pt]},
    draw=rink!75, line width=0.55pt},
  model flow/.style={flow, draw=rgreen, line width=0.85pt}
]
  \colorlet{archhost}{rviolet!13}
  \colorlet{archsandbox}{rgreen!15}
  \colorlet{archexternal}{black!8}
  \colorlet{archdata}{rblue!13}
  \newcommand{\archdetail}[1]{{\fontsize{6.6}{7.6}\selectfont\mdseries #1}}

  \node[card=archdata, text width=1.66cm] (input) at (0.98,0)
    {Inputs\\\archdetail{tasks, models}};
  \node[card=archhost, text width=2.05cm] (loader) at (3.40,0)
    {Task loader\\\archdetail{tasks $\times$ models}};
  \node[card=archhost, text width=2.16cm] (collector) at (6.63,0)
    {Collector\\\archdetail{retries, Resume}};
  \node[optional, draw=rviolet!72, text width=2.15cm] (simulator) at (9.48,0)
    {User Simulator\\\archdetail{separate workspace}};
  \node[card=archexternal, text width=1.87cm] (api) at (12.55,0)
    {Model API\\\archdetail{external service}};
  \draw[flow] (input) -- (loader);
  \draw[flow] (loader) -- node[label text, above] {rollouts} (collector);
  \draw[exchange, dashed] (collector) -- (simulator);

  \filldraw[boundary, fill=rbox, draw=rgreen!55] (5.18,-1.13) rectangle (10.35,-2.56);
  \node[label text, anchor=west, font=\fontsize{7.2}{8.0}\selectfont\bfseries, text=rgreen!75!black]
    at (5.34,-1.37) {Docker or remote sandbox};
  \node[card=archhost, text width=2.05cm] (bundle) at (3.40,-1.99)
    {Task Bundle\\\archdetail{snapshot, digests}};
  \node[card=archsandbox, text width=1.80cm] (workspace) at (6.36,-1.99)
    {Workspace\\\archdetail{inputs, tools}};
  \node[card=archsandbox, text width=2.12cm] (native) at (9.09,-1.99)
    {Native Harness\\\archdetail{agent loop}};
  \node[card=archhost, text width=1.99cm] (proxy) at (11.97,-1.99)
    {Request proxy\\\archdetail{when enabled}};
  \node[card=archhost, text width=2.10cm] (stirrup) at (9.09,-3.22)
    {Stirrup\\\archdetail{host agent loop}};

  \draw[flow] (loader) -- (bundle);
  \draw[flow] (bundle) -- node[label text, above] {inputs} (workspace);
  \draw[flow] (collector.south) -- node[label text, right] {start rollout} (6.63,-1.13);
  \draw[exchange] (workspace) -- (native);
  \draw[exchange] (stirrup.west) -| node[label text, pos=0.25, above] {tool I/O} (workspace.south);
  \draw[model flow] (native) -- node[label text, above, text=rgreen!75!black] {calls} (proxy);
  \draw[model flow] (stirrup.east) -| ([xshift=-6mm]proxy.south);
  \draw[model flow] (proxy.east) -- (13.30,-1.99) -- ([xshift=7.5mm]api.south);

  \node[optional, draw=rorange!75, text width=2.15cm] (grader) at (4.92,-3.22)
    {Bench grader\\\archdetail{workspace or host}};
  \node[card=archdata, text width=1.69cm] (local) at (0.98,-4.59)
    {Local logs\\\archdetail{\texttt{roller import}}};
  \node[card=archhost, text width=4.31cm] (convert) at (7.63,-4.59)
    {Trace conversion and audits\\\archdetail{ACTF, SFT windows, eligibility}};
  \node[card=archdata, text width=7.00cm, align=left, inner xsep=5pt] (run) at (7.63,-5.86)
    {Run directory\\[-0.2pt]
     {\fontsize{6.5}{7.4}\selectfont\mdseries\ttfamily trace.jsonl \enspace trace\_ACTF.json \enspace
      lock.json \enspace info.json}\\[-0.2pt]
     {\fontsize{6.5}{7.4}\selectfont\mdseries\ttfamily trace\_SFT.json \enspace sft\_segments/ \enspace score.json}};
  \node[card=archdata, text width=1.94cm] (export) at (12.54,-5.86)
    {Consumers\\\archdetail{Viewer, exports}};

  \draw[flow, dashed] (5.52,-2.56) -- ([xshift=6mm]grader.north);
  \draw[flow, dashed] ([xshift=6mm]grader.south) -- ++(0,-0.19)
    -| ([xshift=-12mm]convert.north);
  \draw[flow] (5.18,-2.42) -- node[label text, below] {session, files} (3.36,-2.42)
    -- (3.36,-4.59) -- (convert.west);
  \draw[flow] (local) -- node[label text, above] {import} (3.36,-4.59);
  \fill[rink!75] (3.36,-4.59) circle[radius=0.65pt];
  \draw[flow] (stirrup.south) -- node[label text, right] {journal} (stirrup.south |- convert.north);
  \draw[flow] ([xshift=6mm]proxy.south) |- node[label text, pos=0.77, above] {request evidence} (convert.east);
  \draw[flow] (convert) -- (run);
  \draw[flow] (run) -- (export);

  \begin{scope}[shift={(0.05,-1.42)}]
    \foreach \offset/\tone/\entry in {
      0/archhost/collector host,-0.38/archsandbox/task workspace,-0.76/archexternal/external} {
      \fill[\tone, rounded corners=1pt] (0,\offset-0.09) rectangle (0.34,\offset+0.09);
      \node[label text, anchor=west] at (0.46,\offset) {\entry};
    }
    \draw[model flow] (0,-1.24) -- (0.34,-1.24);
    \node[label text, anchor=west] at (0.46,-1.24) {model traffic};
    \draw[flow] (0,-1.62) -- (0.34,-1.62);
    \node[label text, anchor=west] at (0.46,-1.62) {data or control};
    \draw[flow, dashed] (0,-2.0) -- (0.34,-2.0);
    \node[label text, anchor=west] at (0.46,-2.0) {optional};
  \end{scope}
\end{tikzpicture}
\caption{Roller architecture. The Collector schedules rollouts against snapshotted inputs and an isolated
workspace. A native Harness runs in that environment. Stirrup keeps its agent loop and journal on the collector
host and sends tool operations to the workspace. Green arrows show the routes through the host proxy when enabled.
Stirrup calls the model API directly by default. Remote Claude Code can reach the proxy through a provider relay.
The remote DeepSeek path supports recording inside the sandbox.
Bench grading uses the live workspace or selected files copied to the collector before teardown. Captured evidence
supports trace conversion, training export, and eligibility audits. Imported sessions reuse the trace converters.}
\label{fig:arch}
\end{figure}
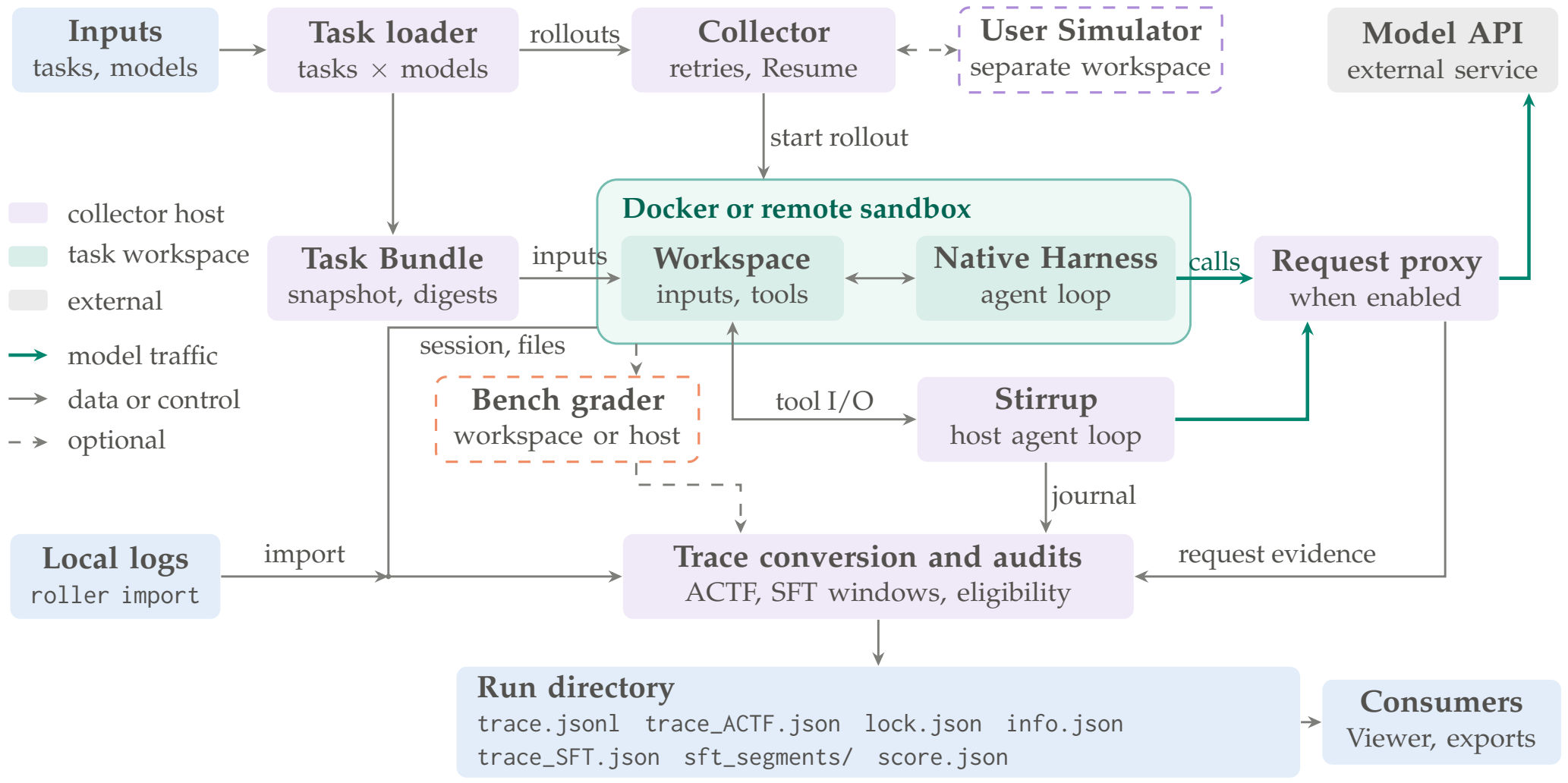

The task loader expands tasks, models and repeated samples into rollout requests (\cref{fig:arch}). The
Collector schedules them in isolated Docker workspaces or remote sandboxes. Most Harnesses run their agent
loop in the task environment. Stirrup keeps its agent loop on the collector and sends tool operations to the
workspace. The adapters expose the Harness's native tools and context management through a common rollout
interface. Interactive adapters also support predefined user turns and, where available, a separate User
Simulator. \Cref{app:system} details the supported Harnesses and execution backends.

\paragraph{Reproducible execution.}
Roller snapshots task inputs and Context files by content before launching a rollout. The run lock binds the
requests to the task bundle digests, model and Harness settings, and resolved container images or sandbox
builds. Resume checks compatibility with the lock and validates candidate samples before reuse. A compatible
run can grow while retaining its existing samples. When the underlying trace is valid, Roller can rebuild a
derived training file without calling the model again (\cref{sec:bundles,sec:lifecycle}).

\paragraph{Capture and grading.}
Each sample records the native session, declared outputs and any captured model traffic. When configured,
the request proxy records outgoing requests and response evidence. Native logs supply events such as tool
execution and delegation, and can provide request evidence for some Harnesses. The capture path depends on
the Harness and backend (\cref{sec:proxy}). Benchmark adapters grade the live workspace or selected files
copied from it before teardown. Roller records grading outcomes, execution status and training eligibility
independently, so a conversion failure leaves a valid benchmark score available (\cref{sec:bench,sec:audits}).

\section{Trajectory Representation and Training Export}
\label{sec:trajectories}

\subsection{Normalized traces}
\label{sec:actf}

Roller converts native sessions to ACTF, a common step representation for visible reasoning, assistant replies,
tool calls and observations, usage and timing. The native session preserves Harness-specific events, while captured
Calls preserve the model requests and responses used to verify context boundaries. \Cref{app:trajectories}
details the artifacts and conversion rules. A linear SFT export projects ACTF to messages and tool schemas
when the observed history supports a single window and the records can be represented in the training format.

\subsection{Context-window reconstruction}
\label{sec:segments}

When capture provides enough evidence, Roller partitions the main agent's calls into append-only windows.
It aligns root Calls to ACTF steps using their responses and starts a new window when a native compaction
boundary occurs, the tool schema changes, or the next request no longer extends the accumulated message
prefix. \Cref{app:segments} describes the matching rules and exceptions for summary requests.

Each segment begins with the first captured prompt in that window. Its \code{target\_start\_index} marks
the first position eligible for assistant supervision. In \cref{fig:teaser}, the second window begins with
the summary $\Sigma$ and the replayed exchange $(A_2,O_2)$. Supervision resumes at $A_3$, so $A_2$ contributes
to the loss only in the first window. Roller writes segmented records to a separate file and converts the
boundaries to per-message loss masks for XTuner. A reset without sufficient captured context makes the
training export unavailable.

\subsection{Sub-agent attribution and training eligibility}
\label{sec:capture}

\paragraph{Sub-agent attribution.}
A captured Call contains a model request and response. Roller assigns it to a root or child participant using
native session or response identifiers. It records delegation, delivery, and the appearance of a result in a
later prompt as separate relations, each supported by native evidence. The main training record contains only
the root participant's steps. Child calls retain their own context (\cref{app:capture,fig:capture}).

\paragraph{Audits.}
\label{sec:audits}
Roller checks visible reasoning, the controls in captured requests, completion evidence and the model name
declared by the provider. Explicit reasoning conflicts, token-limit stops and other recorded exclusions
prevent a sample from entering the training aggregate. An unverified reasoning level can coexist with an
eligible record. Rollout status and any benchmark score are stored separately from training eligibility
(\cref{tab:eligibility,app:audits}).

\section{Experiments}
\label{sec:experiments}

We evaluate reasoning-level verification, benchmark grading, and the use of collected trajectories for
supervised fine-tuning. The runs took place in September and October 2026. \Cref{app:experiments} provides
the experimental configuration and supplementary results.

\subsection{Reasoning-level verification}
\label{sec:exp-thinking}

\begin{figure}[!htbp]
\centering
\includegraphics[width=0.75\columnwidth]{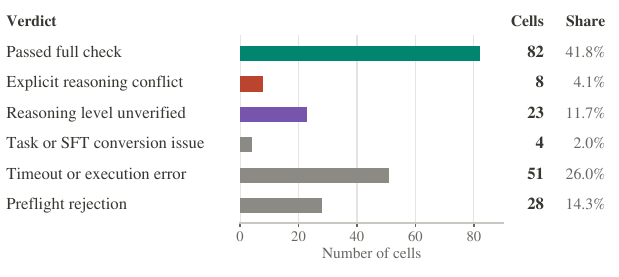}
\caption{Verdicts for the 196 main-matrix cells measured at Roller commit \code{af681e7}. A pass requires
a verified reasoning level in the outgoing request. All eight explicit conflicts were excluded from the
SFT aggregate. \Cref{tab:thinking} includes the smoke and full-wire control runs.}
\label{fig:thinking}
\end{figure}

We tested seven Harnesses, seven models and four reasoning levels on a deterministic write-and-read task.
A full pass required task success, a verified reasoning level and an eligible training record. Of the 196
main-matrix cells, 82 passed, 23 had an unverified level, and eight showed explicit reasoning conflicts
(\cref{fig:thinking}). All eight conflicts were excluded from the SFT aggregate. In one case, the run
requested \code{low} for DeepSeek V4 Flash and the three outgoing OpenClaw requests carried \code{high}.
The marker task still passed. The main matrix produced 106 eligible training rows, illustrating that
eligibility and verified reasoning measure different properties (\cref{tab:thinking}).

\subsection{Benchmark runs}
\label{sec:exp-bench}

On QwenClawBench, each model ran three OpenClaw rollouts on each of 100 tasks with GLM-5.2 as judge.
GLM-5.2 scored 59.26 with 282 valid rollouts out of 300. DeepSeek V4 Flash scored 46.52 with 245 valid
rollouts. On SWE-Bench Pro, DeepSeek V4 Pro resolved 55 of the first 100 public tasks, with 99 valid
results. This run regraded the first 71 patches after a grader fix and included selective High-effort
retries. The headline scores count invalid or unscored results as zero. \Cref{tab:benchrepro}
reports both scoring protocols and gives the published reference scores in its notes.

\begin{table}[!htbp]
\caption{\textbf{Benchmark results collected with Roller.}
The rollouts used OpenClaw \citep{openclaw}. The headline score uses the
full rollout or task denominator shown below, with invalid or unscored results contributing zero.
Valid counts are the completions reported in the original runs. Auto and LLM are the archived QwenClawBench
grader subscores. Bars use the 0--100 scale within QwenClawBench.}
\label{tab:benchrepro}
\vskip 0.06in
\centering
\scriptsize
\setlength{\tabcolsep}{3pt}
\begin{tabularx}{\linewidth}{@{}l >{\raggedright\arraybackslash}X rrccc@{}}
\toprule
& & \multicolumn{2}{c}{\textbf{Evaluation coverage}} & \multicolumn{3}{c}{\textbf{Grader results}} \\
\cmidrule(lr){3-4}\cmidrule(lr){5-7}
\tablehead
\textbf{Benchmark} & \textbf{Model} & \textbf{Denominator} & \textbf{Valid} &
\textbf{Score}\up & \textbf{Auto}\up & \textbf{LLM}\up \\
\midrule
QwenClawBench\tablesource{qwenclawbench} & GLM-5.2\tablesource{glm52} & 300 rollouts & 282 (94.0\%) &
  \resultbar{59.26}{100}{59.26} & \resultbar{72.50}{100}{72.50} & \resultbar{69.10}{100}{69.10} \\
QwenClawBench\tablesource{qwenclawbench} & DeepSeek V4 Flash\tablesource{deepseekv4flash} & 300 rollouts & 245 (81.7\%) &
  \resultbar{46.52}{100}{46.52} & \resultbar{60.99}{100}{60.99} & \resultbar{55.84}{100}{55.84} \\
\addlinespace[2pt]
SWE-Bench Pro\tablesource{deng2025swebenchpro} & DeepSeek V4 Pro\tablesource{deepseekv4pro} & 100 tasks & 99 (99\%) & 55.00 & \na & \na \\
\bottomrule
\end{tabularx}
\par\smallskip
\begin{minipage}{\linewidth}
\raggedright\scriptsize
\textit{Published context.} QwenClawBench reports 58.7 for GLM-5.1 and 50.7 for DeepSeek-V3.2-Thinking
under a Claude judge. Roller used GLM-5.2 as judge and evaluated the models listed above.
The published DeepSeek V4 Pro scores for SWE-Bench Pro are 52.1, 54.4 and 55.4 across three reasoning
settings on all 731 public tasks. Roller's 55/100 is cumulative on the first 100, after regrading
and selective retries at High.
\end{minipage}
\end{table}

\subsection{Collection at scale}
\label{sec:exp-scale}

Two teacher runs with GLM-5.3-Flash on OpenClaw scheduled 1,400 animation and HTML slide tasks. When the runs
were stopped manually, Roller cancelled 682 queued rollouts and finalized all 718 that had started. Of these,
700 produced an ACTF trace and 642 saved a deliverable. The overlapping runs reached 72 concurrent
containers. \Cref{tab:scale} gives the per-suite coverage, lengths and failure counts.

\subsection{Training on collected trajectories}
\label{sec:exp-sft}

We exported the teacher trajectories as linear SFT records and fine-tuned Qwen3.5-35B-A3B with XTuner.
We evaluated each condition on 20 held-out tasks per suite using the same Harness and grader. The teacher's
reasoning lengths had a long tail. The median was 182 characters, while the
97th percentile was 59,878 characters (\cref{fig:reasoning}). We therefore varied which reasoning spans
entered the training record, keeping the surrounding content and tool calls fixed.

On held-out slides, blanking reasoning above 2,000 characters increased the grader score from 9.40 to 9.62,
with all 20 tasks reported complete and all 20 deliverables saved. Blanking all reasoning scored 9.70 and
saved all 20 deliverables, with seven tasks reported complete. On held-out animation, the 12,000-character
and all-blank conditions both scored 10.00 and saved all 20 deliverables. Their completion counts were
20 and three, respectively. These outcomes show why the evaluation reports grader score, deliverable
presence and Harness completion separately (\cref{tab:sft}).

\begin{table}[!htbp]
\caption{\textbf{Held-out evaluation before and after SFT.}
Qwen3.5-35B-A3B \citep{qwen35a3b} was trained on GLM-5.3-Flash trajectories \citep{glm53flash}.
Each suite has 20 held-out tasks. The three SFT arms use the same supervised content and tool calls
(about 14M tokens) and vary how they retain \code{reasoning\_content}.}
\label{tab:sft}
\vskip 0.06in
\centering
\scriptsize
\setlength{\tabcolsep}{3pt}
\begin{tabularx}{0.91\linewidth}{@{}>{\raggedright\arraybackslash}X ccc ccc@{}}
\toprule
& \multicolumn{3}{c}{\textbf{Slides}} & \multicolumn{3}{c}{\textbf{Animation}} \\
\cmidrule(lr){2-4}\cmidrule(lr){5-7}
\tablehead
\textbf{Training condition} &
\textbf{Done} & \textbf{Written} & \textbf{Score}\up &
\textbf{Done} & \textbf{Written} & \textbf{Score}\up \\
\midrule
Base model, no SFT &
  \resultbar[raqua]{20}{20}{20/20} & 20/20 & \resultbar{9.40}{10}{9.40} &
  \resultbar[raqua]{20}{20}{20/20} & 20/20 & \resultbar{9.89}{10}{9.89} \\
SFT, blank above 12k chars &
  \resultbar[raqua]{20}{20}{20/20} & 19/20 & \resultbar{9.28}{10}{9.28} &
  \resultbar[raqua]{20}{20}{20/20} & 20/20 & \resultbar{10.00}{10}{10.00} \\
SFT, blank above 2k chars &
  \resultbar[raqua]{20}{20}{20/20} & 20/20 & \resultbar{9.62}{10}{9.62} &
  \resultbar[raqua]{20}{20}{20/20} & 19/20 & \resultbar{9.50}{10}{9.50} \\
SFT, all reasoning blanked &
  \resultbar[raqua]{7}{20}{7/20} & 20/20 & \resultbar{9.70}{10}{9.70} &
  \resultbar[raqua]{3}{20}{3/20} & 20/20 & \resultbar{10.00}{10}{10.00} \\
\midrule
\textit{Teacher reference} & \textit{181/181} & \textit{181/181} & \textit{9.68} & \textit{401/401} & \textit{401/401} & \textit{9.98} \\
\bottomrule
\end{tabularx}
\par\smallskip
\begin{minipage}{0.91\linewidth}
\raggedright\scriptsize
Done counts successes reported by the Harness. Written counts saved deliverables.
Bars show Done out of 20 and the composite grader score on a 0--10 scale.
The teacher reference is measured on the training tasks.
\end{minipage}
\end{table}

\FloatBarrier
\section{Conclusion}
\label{sec:conclusion}

Roller links each agent trajectory to its inputs, execution environment and native session evidence.
When captured requests support reconstruction, it uses them to recover the context for each supervised
response and exports training windows with explicit supervision boundaries.

The reasoning audit excluded eight explicit conflicts in a 196-cell matrix, including a rollout that
completed its task. Two fine-tuning conditions attained the same animation score and saved all
deliverables, yet their Harness completion counts were 20 and three. These results support separate
reporting of grader scores, execution status and training eligibility.

Captured requests establish what was sent to the service. The provider's effective inference context
requires separate evidence. The fine-tuning study covers one teacher--student pair and two task suites.
The effect of reasoning filtering across other models and tasks remains open.

{
    \small
    \bibliography{refs}
    \bibliographystyle{valenteam}
}

\clearpage
\appendix
\section{Contributions}
\label{sec:contributors}

Ziyu Liu\textsuperscript{1},
Penghui Yang\textsuperscript{1},
Shengyuan Ding\textsuperscript{2},
Yizhuo Li\textsuperscript{1},
Qiyang Zhang\textsuperscript{3},
Yuhang Zang\textsuperscript{\dag}.

\vspace{4pt}
{\small \textsuperscript{1}Shanghai Jiao Tong University \quad
\textsuperscript{2}Fudan University\\
\textsuperscript{3}Sun Yat-sen University}

\vspace{8pt}
\textsuperscript{\dag} Corresponding author\\
\textbf{Contact} \enspace Ziyu Liu,
\href{mailto:liuziyu77@sjtu.edu.cn}{\nolinkurl{liuziyu77@sjtu.edu.cn}}

\subsection{Acknowledgments}

We thank Yini Huang, Qianli Ma, and Di Yu (Orion-DH) for their contributions to Roller.

\section{System and Execution Details}
\label{app:system}

\subsection{Tasks and Task Bundles}
\label{sec:bundles}

A task is a JSON object under the schema \code{roller.task/v1}. Only an identifier and a prompt are required.
Optional fields declare predefined user turns (\code{messages}), a User Simulator, input files, Context files, an
output directory, environment variables, setup commands and a per-task reasoning level.

Before any container starts, \roller{} scans every declared input and Context file. It rejects symbolic links,
special files and paths that escape their declared root, records size, mode and SHA-256 for each file, and checks
declared digests. It then copies each source into a content-addressed input tree, hashing the bytes again during
the copy so that a file changed after the scan is caught. The snapshot is shared by all models that run the task,
and identical input trees within a run are stored once. Three digests summarize a task:
$\digest_{\text{input}}$ over the input tree, $\digest_{\text{ctx}}$ over the Context manifest, and
$\digest_{\text{bundle}}$ over the declaration together with both. These digests, not the files, travel into the
sample metadata and the run lock.

Context differs from ordinary input in that each Harness discovers it at a native location. A \code{role} and a
\code{target} let \roller{} place a file where the agent will find it: \file{AGENTS.md} in the Codex workspace, a
skill under \file{\$CLAUDE\_CONFIG\_DIR/skills/} for Claude Code, \file{USER.md} and \file{MEMORY.md} in both the
OpenClaw workspace and \file{/root}. A capture policy decides what the run keeps. Under \code{full} the source is
stored in the run by content; under \code{hash} only the digest is kept, which suits private user or memory files;
under \code{none} nothing is recorded, and the task gives up safe Resume because \roller{} cannot tell whether the
Context changed.

After the run, \roller{} walks the declared \code{output\_dir}, copies regular files and empty directories into
\file{results/}, and records path, size, MIME type and SHA-256 per file. A symbolic link or special file in the
output is not copied, and its presence turns the sample status to \code{error}, so an output that left the
expected shape is never reported as a success. Generated \file{node\_modules/} trees are skipped.

\subsection{Harness adapters}
\label{sec:harness}

Each Harness adapter implements one interface. A non-interactive adapter exposes \code{run(request)}. An
interactive adapter opens a session with \code{start}, delivers each user turn with \code{send} and closes with
\code{finish}, which exports the native session, collects outputs and removes the container. Multi-turn tasks,
the User Simulator and access to the live workspace for grading all build on the interactive form.

\Cref{tab:harnesses} lists Roller's eight Harness adapters. All support Docker workspaces. Claude Code and
Stirrup support remote sandboxes, and DeepSeek supports them for WildClawBench. Containerized Harnesses use
an isolated home for their sessions and configuration. Stirrup writes its journal in the collector and
executes tools in the task workspace.
Each adapter exposes the selected Harness's tool interface. Flags such as \flag{hermes-toolset} and
\flag{opencode-deny-tool} select tools and are recorded in the run configuration. Sub-agent capture follows
each Harness's capabilities (\cref{sec:capture}). Rollout setup disables OpenClaw's background heartbeat and
configures transcript capture to retain the task text.

\begin{table}[tbp]
\caption{Harness capabilities in Roller. Pins come from the reference images or SDK dependencies.
All eight Harnesses support Docker workspaces. An n/a entry in the remote column indicates Docker-only support.
Stirrup runs its agent loop on the collector. The model API column gives the standard protocols.}
\label{tab:harnesses}
\vskip 0.06in
\centering
\scriptsize
\setlength{\tabcolsep}{3pt}
\begin{tabularx}{\linewidth}{@{}l l >{\raggedright\arraybackslash}X lcc@{}}
\toprule
\tablehead
\textbf{Harness} & \textbf{Pinned release} & \textbf{Model API} & \textbf{Remote sandbox} &
  \shortstack{\textbf{User}\\\textbf{Simulator}} & \shortstack{\textbf{Sub-agent}\\\textbf{capture}} \\
\midrule
OpenClaw\tablesource{openclaw}    & 2026.11.10 & Responses / Chat & \na & \yes & \yes \\
Codex\tablesource{codexcli}       & 0.172.0    & Responses & \na & \yes & \yes \\
Claude Code\tablesource{claudecode} & 2.1.332    & Anthropic Messages & Supported & \yes & \yes \\
Hermes\tablesource{hermesagent}      & \code{8fc037444731} & Chat Completions & \na & \yes & \yes \\
OpenCode\tablesource{opencode}    & 1.19.6     & Chat Completions & \na & \yes & \yes \\
DeepSeek\tablesource{deepseekharness}    & 0.2.2rc2   & DeepSeek Chat & WildClawBench only & \yes & \no \\
Pi Agent\tablesource{piagent}    & 1.6.1      & Chat Completions & \na & \yes & \no \\
Stirrup\tablesource{stirrup}     & 0.2.2      & Chat Completions & Supported & \no & \no \\
\bottomrule
\end{tabularx}
\end{table}

For Hermes, OpenCode, DeepSeek and Pi, the adapter hooks the runtime's own event stream and writes a versioned
event log: serialized provider requests, native message and part updates, usage and turn boundaries, recorded
before the runtime compresses its context. For OpenClaw, \roller{} reads the transcript through
the runtime's public SDK by exact session identity. For Codex and Claude Code it copies the native session files,
and for Claude Code it merges the main session with its \file{subagents/} logs by timestamp and deduplicates by the
native event identifier. Stirrup supplies its native journal. \file{trace.jsonl} preserves each Harness's event
format and ACTF normalizes the steps (\cref{sec:actf}).

\subsection{Request proxy}
\label{sec:proxy}

The reverse proxy can record and adapt traffic between the Harness and the model API. Its use depends on the
Harness and the run configuration. For example, \flag{no-keep-thinking} strips earlier reasoning from OpenClaw
requests at the proxy. The Claude Code proxy removes a transport-only beta query and header and records the names
of removed headers. The wire log omits authorization headers. Roller removes its own correlation headers
(\code{x-roller-*}) before forwarding and links response metadata, including declared model names and finish
reasons, to requests by sequence number.

\flag{wire-trace} controls request recording. In the default \code{actf} mode, the proxy, when used, records
the tool schemas and other fields needed for conversion. The temporary wire trace is pruned after conversion.
Codex retains its captured tool definitions in \file{tools\_capture.jsonl}. The \code{full} mode saves complete
request bodies, so log size can grow roughly quadratically when each request repeats the accumulated history.
The \code{off} mode disables ordinary wire recording. OpenClaw can still start a proxy for request
compatibility, including reasoning-history rewrites. Stirrup calls the model API directly by default and
uses the host proxy with \code{full} recording or \flag{capture-calls}. Tool definitions can also come from native
records or built-in schemas. Claude Code can derive a permissive schema from observed tool calls without a proxy.
When the converter lacks a needed schema, it records why SFT export is unavailable.

\subsection{Lifecycle, run lock and Resume}
\label{sec:lifecycle}

\Cref{alg:rollout} gives the life of one rollout. A collection run writes \file{lock.json} before the first rollout.
The lock records the requests and each task's $\digest_{\text{bundle}}$, together with the model and Harness
configuration, the resolved image IDs or sandbox build, and execution settings such as the timeout, reasoning
level and capture options. Credentials are
replaced by \code{<redacted>}, URL user information and fragments are removed, and sensitive query values are
masked while keys such as \code{api-version} stay comparable. Scheduling settings such as \flag{parallel} and
\flag{retries} are left out, because they do not change a finished sample.

\begin{algorithm}[tbp]
\caption{One rollout under \flag{resume} and \flag{retries}~$N$}
\label{alg:rollout}
\begin{algorithmic}
\small
\STATE {\bfseries Input:} request $r=(\tau,m,h,i)$, run lock $\Lambda$, old sample $s$ (may be empty)
\IF{$s.\text{status}=\text{ok}$ \AND $\textsc{Reusable}(s, r, \Lambda)$}
  \STATE mark $s$ \code{skipped}; \textbf{return} $s$
\ENDIF
\FOR{$a = 1$ {\bfseries to} $N+1$}
  \STATE clear runtime files of attempt $a-1$
  \STATE start the task environment from the resolved image or build and stage the task bundle
  \STATE run turns (predefined, or relayed from the User Simulator)
  \STATE grade the live workspace or copied files, if the task has a grader
  \STATE collect the session and outputs and release the task environment
  \IF{status $=$ ok} \STATE {\bfseries break} \ENDIF
\ENDFOR
\STATE finalize the last attempt, converting available traces and running audits
\STATE write \file{info.json} with every attempt and \textbf{return} the sample
\end{algorithmic}
\end{algorithm}

Resume checks the new invocation against the stored lock before launching any rollout. Compatible runs may
add tasks or models, change existing requests' positions, or use another patch release in the same minor series.
Changes to locked inputs or execution settings for existing requests stop the run with a report of the differences.
For each candidate sample, Resume checks its identity and bundle provenance. When the full artifacts are retained,
it validates the available traces, required capture files and recorded output paths and sizes.
\flag{resume-verify-outputs} also compares output SHA-256 digests. If the evidence is valid, Resume can
rebuild a linear SFT file from ACTF or a segment file from captured Calls without rerunning the model.
Samples retained with \flag{training-files-only} use the compact metadata path for reuse. Samples that fail
integrity checks generally rerun. A run created before locks existed is migrated by rerunning every sample
once. A durable marker blocks reuse until the rerun and run-level aggregation finish.

Each rollout inspects the image ID again and starts its container from that ID, so the provenance names the image
that ran even if a tag moves during the run. The run-level \file{info.json} is written atomically after
the first result, after every 25 further results, and whenever 30 seconds pass between completions. Before
the first rollout, \roller{} checks that every host variable and API key the run reads is set. During the run it
prints progress and an estimated finish time, logs a heartbeat when 60 seconds pass without a completion, and appends
its log to \file{collect.log}; after 10 consecutive errors
(\flag{max-consecutive-errors}) it stops dispatching new rollouts, so a broken endpoint does not burn through the
whole task list. On a catchable interrupt, queued rollouts are cancelled, running ones finish normally, and a partial
\file{sft.jsonl} is built from the finished samples. Containers carry labels that name the collector process that
owns them, and \code{roller clean} removes those left behind by a collector that died.

\subsection{Execution backends}
\begin{figure}[tbp]
\centering
\begin{tikzpicture}[
  x=1cm, y=1cm,
  header/.style={fill=#1, rounded corners=3pt, minimum height=0.53cm, inner sep=2pt,
    align=center, font=\fontsize{8.0}{9.0}\selectfont\bfseries, text=rink},
  step/.style={fill=#1, rounded corners=2.5pt, minimum height=0.40cm, inner sep=2pt,
    align=center, font=\fontsize{7.1}{8.0}\selectfont, text=rink},
  phase/.style={anchor=west, inner sep=0pt, font=\fontsize{7.6}{8.6}\selectfont\bfseries, text=rink},
  label text/.style={font=\fontsize{7.0}{8.0}\selectfont, text=rink, inner sep=2pt},
  flow/.style={-{Stealth[length=3.5pt,width=3.2pt]}, draw=rink!75, line width=0.55pt},
  traffic/.style={{Stealth[length=3.5pt,width=3.2pt]}-{Stealth[length=3.5pt,width=3.2pt]},
    draw=rgreen, line width=0.85pt},
  lifeline/.style={draw=rink!35, line width=0.45pt, dash pattern=on 2pt off 2pt}
]
  \colorlet{sandboxhost}{rviolet!13}
  \colorlet{sandboxwork}{rgreen!15}
  \colorlet{sandboxrelay}{rblue!13}
  \colorlet{sandboxapi}{black!8}
  \def\sandboxx{2.36}
  \def\relayx{5.45}
  \def\hostx{8.80}
  \def\modelx{12.02}

  \fill[black!2, rounded corners=4pt] (0,-0.43) rectangle (13.56,-3.44);
  \fill[rgreen!4, rounded corners=4pt] (0,-3.56) rectangle (13.56,-5.44);
  \fill[black!2, rounded corners=4pt] (0,-5.56) rectangle (13.56,-8.34);

  \node[header=sandboxwork, text width=2.72cm] at (\sandboxx,0) {Remote sandbox};
  \node[header=sandboxrelay, text width=2.28cm] at (\relayx,0) {Provider relay};
  \node[header=sandboxhost, text width=2.55cm] at (\hostx,0) {Collector host};
  \node[header=sandboxapi, text width=1.95cm] at (\modelx,0) {Model API};
  \foreach \x in {\sandboxx,\relayx,\hostx,\modelx}
    \draw[lifeline] (\x,-0.27) -- (\x,-8.20);

  \node[phase] at (0.18,-0.69) {Provision};
  \node[step=sandboxhost, text width=2.52cm] at (\hostx,-1.00)
    {Pin build\\start host proxy};
  \draw[flow] (\hostx,-1.59) -- node[label text, above] {Create sandbox from pinned build} (\sandboxx,-1.59);
  \node[step=sandboxhost, text width=2.52cm] (receipt) at (\hostx,-2.08)
    {Save allocation receipt};
  \draw[flow] (\sandboxx,-2.08) -- node[label text, above] {Report instance ID} (receipt.west);
  \draw[flow] (\hostx,-2.65) -- node[label text, above] {Copy inputs and Context} (\sandboxx,-2.65);
  \draw[flow] (\hostx,-3.16) -- node[label text, above] {Launch CLI with relay URL} (\sandboxx,-3.16);

  \node[phase] at (0.18,-3.83) {Run and capture};
  \node[step=sandboxwork, text width=2.22cm, minimum height=0.56cm] (claude) at (\sandboxx,-4.40)
    {Claude Code CLI\\agent loop};
  \node[step=sandboxrelay, text width=2.16cm, minimum height=0.56cm] (relay) at (\relayx,-4.40)
    {HTTP relay\\via sandbox API};
  \node[step=sandboxhost, text width=2.52cm, minimum height=0.56cm] (proxy) at (\hostx,-4.40)
    {Request proxy\\forward to model API};
  \node[step=sandboxapi, text width=1.95cm, minimum height=0.56cm] (model) at (\modelx,-4.40)
    {Model inference};
  \draw[traffic] (claude) -- (relay);
  \draw[traffic] (relay) -- (proxy);
  \draw[traffic] (proxy) -- (model);
  \node[step=sandboxwork, text width=2.24cm] (session) at (\sandboxx,-5.09) {Session and files};
  \node[step=sandboxhost, text width=2.52cm] (capture) at (\hostx,-5.09) {Wire trace and Calls};
  \draw[flow] (claude) -- (session);
  \draw[flow] (proxy) -- (capture);

  \node[phase] at (0.18,-5.80) {Finalize};
  \node[step=white, dashed, draw=rorange!75, text width=2.52cm] (grade) at (\hostx,-6.00)
    {Grade on collector};
  \draw[flow, dashed] (\sandboxx,-6.00) -- node[label text, above] {Copy selected grading files} (grade.west);
  \node[step=sandboxhost, text width=2.52cm] (normalize) at (\hostx,-6.63)
    {Check files\\build ACTF};
  \draw[flow] (\sandboxx,-6.63) -- node[label text, above] {Download final session and outputs} (normalize.west);
  \draw[flow] (\hostx,-7.24) -- node[label text, above] {Release sandbox} (\sandboxx,-7.24);
  \node[step=sandboxhost, text width=2.52cm, minimum height=0.56cm] at (\hostx,-7.91)
    {Clear receipt\\audit and finalize};
\end{tikzpicture}
\caption{A Claude Code rollout in a remote sandbox without direct egress. The provider relays model requests
and responses (green) between the CLI and the collector's request proxy. The collector can retain wire evidence
and captured Calls according to the run configuration. When grading runs on the collector, selected files are
copied out while the sandbox is live. On Resume, the collector uses surviving allocation receipts to release
leftover sandboxes.}
\label{fig:sandbox}
\end{figure}
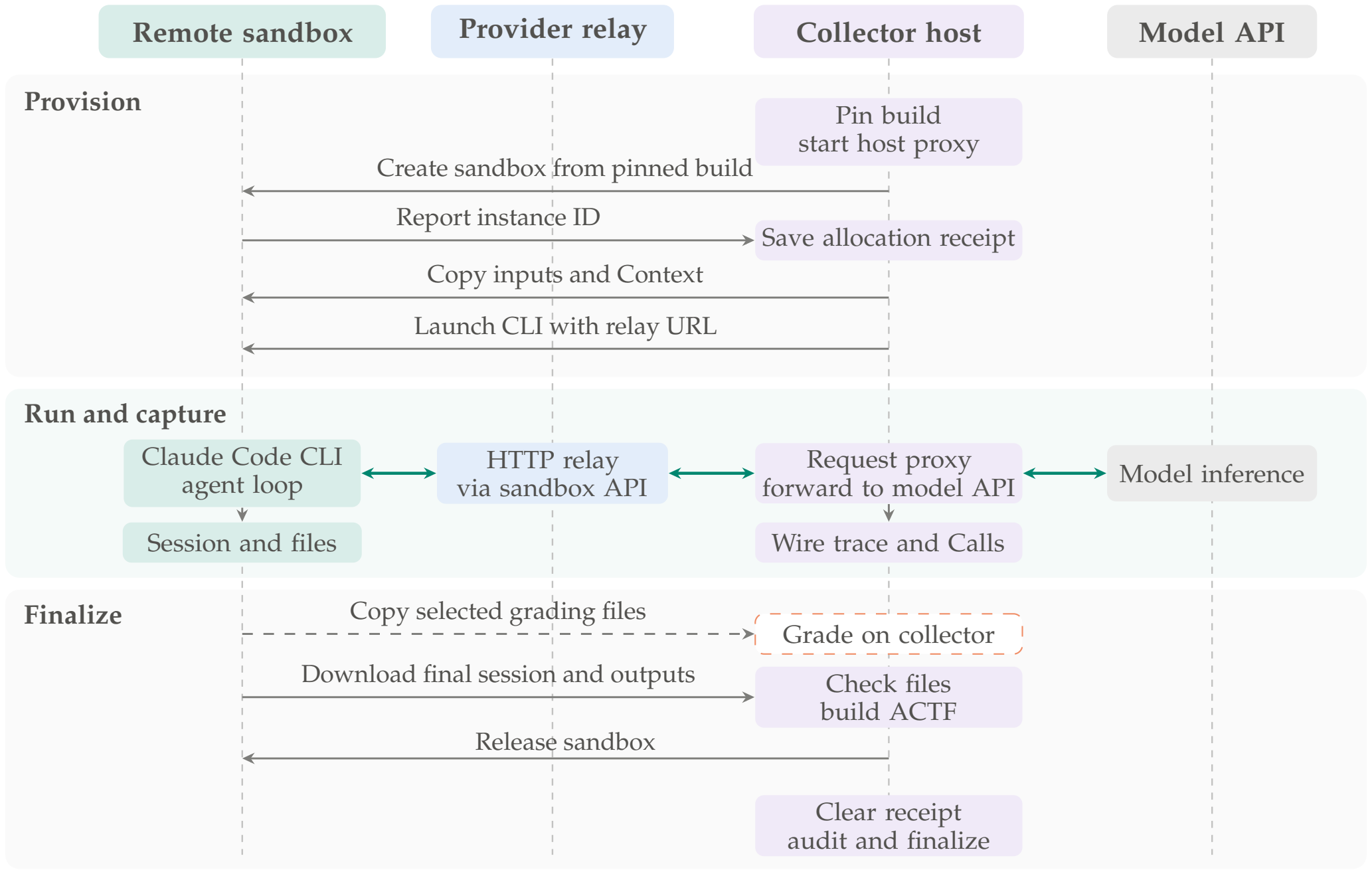

Docker is the default execution backend. Supported Harnesses also run in remote sandboxes, selected with
\flag{sandbox-provider} and a prebuilt environment template. The provider interface resolves a template to a
stable build, creates and removes instances, executes commands, and transfers files. Roller ships an optional
E2B provider and loads additional providers through a Python entry point.

\roller{} pins the provider, template and build in the run configuration and checks the build when each sandbox
starts. It writes an allocation receipt under \file{.sandbox-resources/} as soon as the provider reports the
instance ID. Closing an instance clears its receipt, and Resume uses remaining receipts to release leftover
sandboxes. Task files and Context are copied into the workspace. Output collection checks transferred files
before it records the sample artifacts.

The capture location depends on the Harness. The remote Claude Code CLI uses the collector's request proxy when
capture is enabled. A provider with HTTP relaying can carry that traffic across an environment without direct
egress (\cref{fig:sandbox}). DeepSeek Harness on WildClawBench can capture full requests or Calls inside the
sandbox and transfer the recordings back with the session. Stirrup runs the agent loop and model client on the
collector and uses the remote sandbox for tool execution. Bench grading can run in the task environment or
on selected files copied to the collector before teardown.

\subsection{User Simulator}
\label{sec:simulator}

A task can name a second Harness and model that play the user. A Conversation Manager holds both sessions for the
whole rollout. On each turn it sends the Simulator a fixed instruction, the number of rounds completed, the turn
limit and the full visible transcript. The Simulator must answer with exactly one JSON object,
\code{\{"action":"message","content":...\}} or \code{\{"action":"finish","reason":...\}}. The Manager parses the
answer with a JSON parser and does not look for end markers in free text; a code fence, extra prose or an unknown
action fails the attempt. The Task Agent owns the inputs, Context and output directory, while the Simulator works
in an isolated workspace and learns about progress only from the transcript. After the last allowed turn the
Simulator is asked once more, so that it can confirm completion; an unsent follow-up is recorded in
\code{termination.pending\_instruction}. Both participants keep their own raw trace, ACTF and SFT, and
\file{conversation.json} stores only what both of them could see. The run-level training file includes only the
Task Agent. The seven Harnesses with User Simulator support in \cref{tab:harnesses} can take either role.
All 49 Agent/Simulator pairings among them run through one code path.

\subsection{Viewer and exports}
\label{sec:viewer}

Roller is available on PyPI as \code{roller-agent} under the Apache 2.0 license \citep{roller2026repo}.
\code{roller view} serves a read-only web interface over a run directory. It needs no extra
dependencies, lists the rollouts of one task across models, and shows each step with its reasoning, tool calls, tool
results, tokens and timing; it reads finished samples while a run is still going. \code{roller export} converts a
run to ATIF, the trajectory format of Harbor \citep{harbor}, whose Claude Code and Codex agents can load a trajectory
as session history; to a Hugging Face dataset folder \citep{lhoest2021datasets} with a dataset card, the agent session
of every sample and a table with one row per sample; or to SFT records with an index of the task, model and bench
score of each, where \flag{min-score} keeps only samples at or above a score. The Hugging Face export replaces home
directories, user names, account identifiers and the values of credential variables, and writes a review report of
the matches it could not classify next to the dataset.

\code{roller import} goes the other way. It reads the local session files of Claude Code and Codex, including
compressed Codex sessions, and writes them as a run that the Viewer and every export read. An imported sample records
its title, its project and where its tools ran. Sessions that a developer produced in daily work, outside any
collection, can thus be inspected, filtered and exported with the same tools as collected rollouts. They carry no
Task Bundle or run lock, since \roller{} did not run them, and their samples say so.

\subsection{Benchmark grading before teardown}
\label{sec:bench}

Benchmark graders often need the artifacts an agent leaves in its workspace, such as a generated page or a
patched repository. \code{GradedHarness} wraps an interactive Harness and grades the rollout before
\code{finish} releases its Docker container or sandbox. The same lifecycle supports retries of a grader while
the workspace is live.

A bench adapter converts upstream tasks into \code{roller.task/v1} rows with a \code{roller.bench/v1} binding.
The binding specifies the grader, its arguments, the environment variables it may receive, and any files to stage.
Roller stages references needed by the grader after the agent's final turn. Most graders execute in the task
environment. A binding with \code{collector\_files} copies just the specified workspace files to the collecting
process and grades the copy. VIALS uses this path for the answer file, keeping judge credentials on the collector.

A valid zero is a score. A crash, timeout or missing score from the grader is recorded as an unscored sample
with an error. \file{bench-report.json} groups results by model and Harness. The default aggregate counts
unscored samples as zero in the headline average and also reports the average over scored samples. Adapters
can add benchmark-specific summaries, including repeated-sample metrics. With \flag{resume},
\flag{resume-unscored} reruns the agent and grader for samples whose agent finished but grading failed.
A valid bench score remains available when training conversion fails.

\Cref{tab:benches} lists Roller's seven benchmark adapters. They cover general agent tasks
\citep{wildclawbench,skillsbench,pinchbench,qwenclawbench}, software engineering
\citep{deng2025swebenchpro} and visual reasoning \citep{lau2026vials,roberts2025zerobench}.
\code{roller bench run} loads a YAML or JSON file supplied through \flag{config}, with optional overlays
through \flag{extra-config}. \code{roller collect --bench} accepts benchmark options as command-line flags.
Each adapter documents upstream settings
and environment variables. In \cref{fig:cases}, panel (b) shows how a missing WildClawBench setting caused
all 131 searches to fail even though the graders could still score the runs.

\begin{table}[tbp]
\caption{Benchmark adapters in Roller. VIALS grades copied answers on the collector. The other
adapters grade in the task environment. Task counts refer to the listed upstream splits or versions.}
\label{tab:benches}
\vskip 0.06in
\centering
\scriptsize
\setlength{\tabcolsep}{3.2pt}
\renewcommand{\arraystretch}{1.14}
\begin{tabularx}{\linewidth}{@{}l >{\raggedright\arraybackslash}p{0.18\linewidth}
  l >{\raggedright\arraybackslash}X@{}}
\toprule
\tablehead
\textbf{Benchmark} & \textbf{Task scope} & \textbf{Default Harness} & \textbf{Scoring protocol} \\
\midrule
WildClawBench\tablesource{wildclawbench} & 60 & OpenClaw\tablesource{openclaw} &
  Per-task graders, some with an LLM judge \\
SkillsBench\tablesource{skillsbench} & 87 (v1.1) & OpenCode\tablesource{opencode} &
  Official tests, with skills or without skills \\
PinchBench\tablesource{pinchbench} & Upstream manifest & OpenClaw\tablesource{openclaw} &
  Automated, LLM or hybrid graders \\
QwenClawBench\tablesource{qwenclawbench} & 100 & OpenClaw\tablesource{openclaw} &
  Automated and LLM graders, with repeated rollouts \\
SWE-Bench Pro\tablesource{deng2025swebenchpro} & 731 (public) & OpenClaw\tablesource{openclaw} &
  Hidden tests in the official instance environments \\
VIALS\tablesource{lau2026vials} & 161 & Stirrup\tablesource{stirrup} &
  LLM judge on the copied answer file \\
ZeroBench\tablesource{roberts2025zerobench} & 100 main or 334 subquestions & Claude Code\tablesource{claudecode} &
  Exact match on the final brace-delimited answer \\
\bottomrule
\end{tabularx}
\end{table}

\FloatBarrier
\section{Capture and Export Details}
\label{app:trajectories}

\Cref{tab:layers} summarizes the evidence retained at each stage of collection and conversion.

\begin{table}[tbp]
\caption{Trace evidence and retention in Roller. In \flag{training-files-only} mode, successful samples
with training files and a valid grade where applicable retain their training records, sample metadata, any score,
and any requested token IDs. The diagnostic traces below are pruned from those samples after conversion and auditing.}
\label{tab:layers}
\vskip 0.06in
\centering
\scriptsize
\setlength{\tabcolsep}{3pt}
\renewcommand{\arraystretch}{1.12}
\begin{tabularx}{\linewidth}{@{}l >{\raggedright\arraybackslash}p{0.23\linewidth}
  >{\raggedright\arraybackslash}X >{\raggedright\arraybackslash}p{0.29\linewidth}@{}}
\toprule
\tablehead
\textbf{Layer} & \textbf{Artifact} & \textbf{Evidence} & \textbf{Availability} \\
\midrule
Session & \file{trace.jsonl} & Native events, event order and usage where recorded by the Harness &
  Collected or imported when a native session is available \\
Model traffic & \file{wire\_trace\allowbreak.jsonl}\par Captured Calls &
  Request bodies forwarded by the proxy, with per-call responses and attribution in Call captures &
  \flag{wire-trace full} retains wire requests. \flag{capture-calls} records Calls \\
Normalized trace & \file{trace\_ACTF\allowbreak.json} &
  A common step schema for reasoning, tools, observations, tokens and timing &
  Written when the native session can be converted \\
\bottomrule
\end{tabularx}
\end{table}

\subsection{ACTF: one step schema across Harnesses}
\label{app:actf}

ACTF is the analysis trace shared by all Harnesses. Each assistant message of the main agent becomes one step. A
step holds the user content of that turn, the assistant's visible reasoning, visible reply and tool calls, the
observations returned for those calls, a metric block (\code{prompt\_tokens\_len}, \code{completion\_tokens\_len},
cache reads and writes), start and end times and the stop reason. The top level holds the system prompt, the
tool schemas and the trajectory timing.

Each Harness has its own converter, because the native records differ in where facts live. Claude Code streams one
assistant message as several native events with the same message identifier; the converter merges them, writes
plaintext thinking into \code{reasoning\_content}, keeps signed or redacted thinking only in the raw trace, and
pairs each \code{tool\_use} with its later \code{tool\_result}. Codex keeps the system content in base
instructions, developer messages and an \code{<environment\_context>} block, and its tool definitions, including
namespaces and custom grammars, are taken from the captured Responses requests. OpenClaw
takes its assembled system prompt from the first captured context-assembly record and its tools from the first
non-empty schema on the wire. Usage accounting follows each Harness's own counters. Where a Harness records a
delivered reply twice, \roller{} removes the duplicate only when the text, the source and the usage all match within
the same turn, so a genuinely repeated reply is kept.

ACTF freezes the system prompt and tools once at the top level. It does not store the request of every step,
which is what \file{wire\_trace.jsonl} under \code{full} is for. When the converter cannot represent a raw trace,
the sample keeps raw and writes the failure under \code{runtime.trajectory\_conversion}, separate from the
collection status; a rollout that ran correctly is still an evaluation result even if its trace cannot be
normalized.

\subsection{Linear SFT records}
\label{app:sft}

The linear SFT record projects ACTF into \code{messages} and \code{tools}. The first message holds the system
prompt. Assistant messages carry \code{reasoning\_content}, \code{content} and \code{tool\_calls}. Run, task,
model, usage and timing metadata stay in \file{info.json}.

The exporter writes a record only when all of the following hold: the trace has a user message and a non-empty
assistant message; every tool call has a non-empty identifier that is unique in the record; every call has exactly
one result in the same step; every called tool exists in the captured schema; function arguments encode as a JSON
object; and every content block is text or an image with embedded base64 bytes. Results of parallel tool calls are
paired by identifier and emitted in the order of the calls, so the record is canonical, while raw and ACTF keep the
actual completion order. A multimodal message uses OpenAI-style content blocks. Images appear in the user or tool
message that carries them as \code{image\_url} blocks with base64 data URIs. The adapter must materialize images
referenced only by path while the task environment is live.

Anything that fails these checks is recorded, with its reason, as \code{target: sft, status: unavailable}. The
collection status does not change, the raw and ACTF evidence stay, and Resume does not call the model again to
repair an export. Run-level \file{sft.jsonl} aggregates only eligible linear records from successful samples, in
request order, and adds no metadata to them.

\subsection{Compaction segments}
\label{app:segments}

When capture provides enough evidence, \roller{} exports one record per append-only window after compaction.
In \cref{fig:teaser}, the second
window starts with the summary $\Sigma$ and the replayed exchange $(A_2,O_2)$ in its first captured prompt.
Supervision resumes at $A_3$, so $A_2$ contributes to the loss only in the first window. The construction uses
the captured Calls of the root participant (\cref{sec:capture}) for prompts and responses, and native
compaction rows to locate boundaries. It proceeds in three steps.

First, Roller binds root Calls to ACTF main steps in request order by comparing each Call's output with the
next step's reasoning, reply and tool calls. For Codex and OpenClaw, the next native compaction event can
identify an unmatched Call as a summary by recording its response text. Stirrup's journal identifies the Call
whose summary replaced the history, and Roller exports that Call as its own training segment. Other unmatched
Calls or steps make the export unavailable.

Second, Roller groups calls into windows. Call $i{+}1$ continues a window when both calls belong to the same
native context, the tool schemas match, and $x_{i+1}$ extends the current message prefix without inserting
assistant messages. Prefix comparison accounts for replayed tool identifiers and equivalent tool arguments.
The record keeps the identifiers from the newest prompt. Roller labels later window boundaries as
\code{native\_boundary}, \code{observed\_context\_change} or, for a standalone summary, \code{summary\_request}.

Third, each record stores \code{messages}, \code{tools} and \code{target\_start\_index}, the length of its first
actual prompt. Assistant messages at or after that index are targets. User and tool messages between them are
context. \flag{no-keep-thinking} clears visible reasoning from both the recorded prompt and the targets.
Roller stores segment files under \file{sft\_segments/} and aggregates them in \file{sft\_segments.jsonl}.
The linear \file{sft.jsonl} is for loaders that supervise every assistant message. Opaque prompt content such as
encrypted reasoning can make the compacted sample unavailable for export.

\subsection{Sub-agent capture}
\label{app:capture}

\begin{figure}[tbp]
\centering
\begin{tikzpicture}[
  x=1cm, y=1cm,
  entry/.style={fill=#1, rounded corners=2.5pt, minimum height=0.67cm, inner sep=2pt,
    align=center, font=\fontsize{8}{9}\selectfont\bfseries, text=rink},
  track/.style={anchor=west, align=left, inner sep=0pt,
    font=\fontsize{8}{9}\selectfont\bfseries, text=rink},
  label text/.style={font=\fontsize{7}{8}\selectfont, text=rink, inner sep=1.4pt},
  link/.style={-{Stealth[length=3.6pt,width=3pt]}, draw=#1, line width=0.70pt}
]
  \newcommand{\captureDetail}[1]{{\fontsize{6.7}{7.6}\selectfont\mdseries #1}}

  \fill[rgreen!5, rounded corners=4pt] (0,0.54) rectangle (13.54,-0.54);
  \fill[rviolet!5, rounded corners=4pt] (0,-1.00) rectangle (13.54,-2.08);
  \fill[rblue!5, rounded corners=4pt] (0,-2.55) rectangle (13.54,-3.63);

  \node[track] at (0.23,0) {Parent Calls\\\captureDetail{HTTP capture}};
  \node[track] at (0.23,-1.54) {Parent session\\\captureDetail{Claude Code log}};
  \node[track] at (0.23,-3.09) {Child participant\\\captureDetail{identity and Calls}};

  \node[entry=rgreen!15, text width=3.21cm] (parent) at (4.69,0)
    {Call $P_1$\\\captureDetail{\texttt{Agent} tool use \texttt{t7} in response}};
  \node[entry=rgreen!15, text width=3.21cm] (later) at (10.98,0)
    {Call $P_2$\\\captureDetail{\texttt{tool\_result(t7)} in request}};

  \node[entry=rviolet!16, text width=3.21cm] (delegation) at (4.69,-1.54)
    {Delegation event\\\captureDetail{\texttt{tool\_use.id = t7}}};
  \node[entry=rviolet!16, text width=3.21cm] (delivery) at (10.98,-1.54)
    {Terminal tool result\\\captureDetail{result for \texttt{t7}, child ID \texttt{c9}}};

  \node[entry=rblue!16, text width=1.72cm] (child) at (4.69,-3.09)
    {Agent \texttt{c9}\\\captureDetail{native identity}};
  \node[entry=rblue!16, text width=1.32cm] (first) at (7.12,-3.09) {Call $C_1$};
  \node[entry=rblue!16, text width=1.32cm] (last) at (9.18,-3.09) {Call $C_2$};

  \draw[link=rink!65, dashed] (parent) -- node[label text, right] {same response ID} (delegation);
  \draw[link=rorange!85!black] (delegation) -- node[label text, right,
    text=rorange!75!black] {\texttt{spawn}} (child);
  \draw[link=rblue!75!black] (child) -- (first);
  \draw[link=rblue!75!black] (first) -- (last);
  \draw[link=rviolet] (last.east) -| node[label text, pos=0.75, right,
    text=rviolet!70!black] {\texttt{result\_delivered}} (delivery.south);
  \draw[link=rgreen] (delivery) -- node[label text, right, text=rgreen!75!black]
    {\texttt{result\_in\_context}} (later);

  \draw[link=rink!60] (3.05,-3.95) -- node[label text, below] {request order} (12.80,-3.95);
\end{tikzpicture}
\caption{An illustrative foreground delegation in Claude Code. Roller links the captured parent response to
its native session event by response ID. The terminal tool result identifies the child and supplies native
evidence for \code{spawn} and \code{result\_delivered} for tool invocation \code{t7}. To record
\code{result\_in\_context}, Roller checks that a captured parent request contains a tool result with the
matching identifier. A relation whose parent Call cannot be bound uniquely is recorded as \code{unknown}.}
\label{fig:capture}
\end{figure}
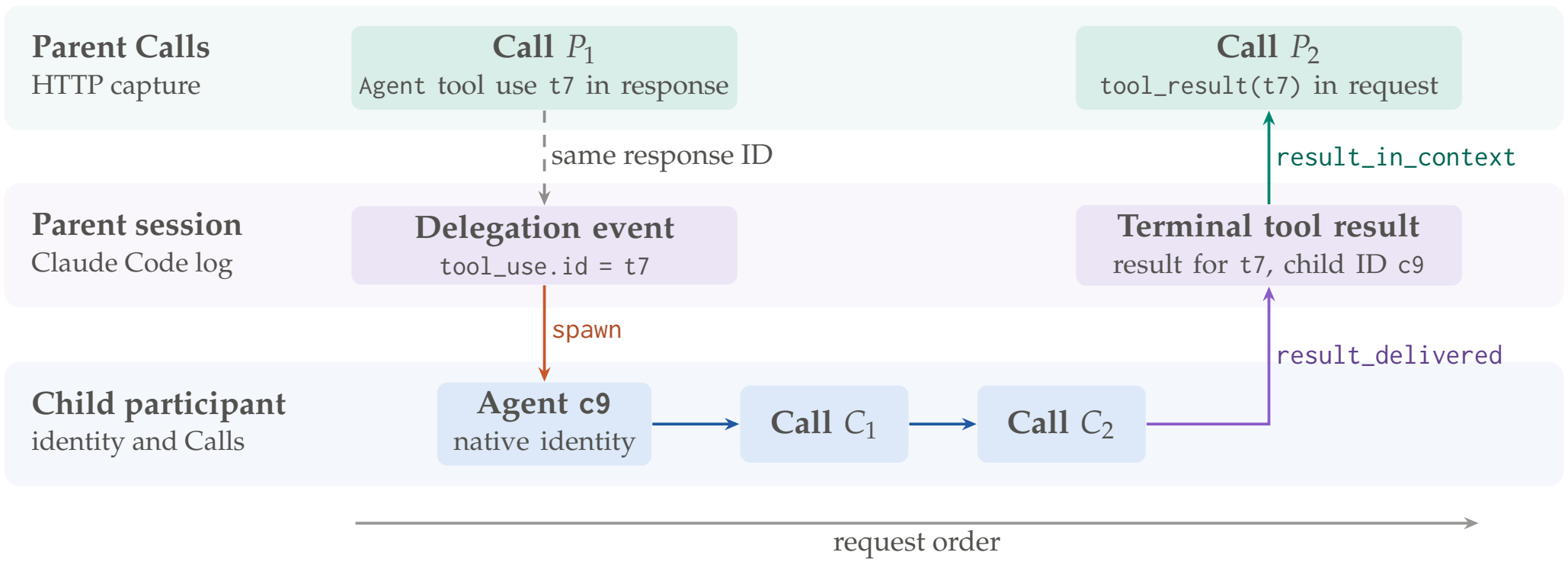

Harnesses delegate work to sub-agents, which issue their own model calls with their own prompts and tools. Under
\flag{capture-calls}, \roller{} records these calls with a protocol built on three objects. A \emph{participant} is
one native executor, identified by an opaque identifier and a role (root or child). A \emph{Call}
(\code{roller.call/v1}) is one HTTP model request and its response: the complete request body as sent, the response
reassembled from the stream, the protocol detected from the endpoint (Anthropic Messages, Chat Completions or
Responses) and the raw response file. A \emph{relation} (\code{roller.relation/v1}) links calls and participants and
carries references to the native file and event that prove it. In \cref{fig:capture}, the terminal result for
tool invocation \code{t7} identifies the child \code{c9}. The captured request for $P_2$ confirms that the
result entered the parent's context. Relation types include
\code{spawn}, \code{result\_delivered}, \code{child\_completed}, \code{child\_failed}, \code{result\_in\_context},
\code{launch\_acknowledged} and \code{launch\_ack\_in\_context}. A background launch acknowledgement is never read as
completion, and a result that arrives as a notification must be bound to the native run or response identifier and
to the notification text in the actual request; matching text alone does not count.

Each Call carries five statuses: whether it finished executing, whether its response reached disk completely,
whether its owner is known, whether the record is complete, and whether it may be exported. A Call is exportable
only when the response ended normally, input and output are complete, the owner is unambiguous, the model matches
and no capture error occurred. A Chat Completions response that ends with a tool-call finish reason but contains no
tool-call identifier or name is marked partial. The capture as a whole can be partial and still contain complete,
exportable Calls.

The protocol splits into a common layer and per-Harness adapters. An adapter returns only native facts:
participants, identity mappings, native relations, observable output identifiers or counts, and a declaration of
its capabilities. The common layer pairs HTTP calls with participants, decides statuses, derives the
context relations and writes the manifest. Identity evidence differs per Harness (\cref{tab:identity}), and the
manifest declares which checks count outputs and which match identifiers. \roller{} never
attributes a call by time, text or request order. Codex, Hermes and OpenCode enable their native sub-agent tools only
under this capture, and the bundled DeepSeek profiles never offer theirs. Under capture, the main ACTF and SFT
contain only the root participant, and a child's messages stay in its own Calls. Resume refuses an older Claude
Code sample whose main trace still mixes in sub-agent steps, instead of reusing it.

\begin{table}[tbp]
\caption{Native evidence for call attribution in Roller. The capture manifest records both the identity key
and the coverage check. Identifier checks reconcile individual outputs. Count checks reconcile totals.}
\label{tab:identity}
\vskip 0.06in
\centering
\scriptsize
\setlength{\tabcolsep}{3pt}
\renewcommand{\arraystretch}{1.1}
\begin{tabularx}{\linewidth}{@{}l >{\raggedright\arraybackslash}X
  >{\raggedright\arraybackslash}p{0.33\linewidth}@{}}
\toprule
\tablehead
\textbf{Harness} & \textbf{Identity evidence} & \textbf{Output coverage} \\
\midrule
Claude Code\tablesource{claudecode} & provider response ID equals the native message ID & per message ID \\
OpenCode\tablesource{opencode} & session ID from \code{chat.headers} equals the native log & completed assistants per session \\
Codex\tablesource{codexcli} & native \code{thread-id} of each request equals the rollout \code{session\_meta} & token-usage updates (count) \\
Hermes\tablesource{hermesagent} & the executing instance's session ID equals the native log & provider response ID \\
OpenClaw\tablesource{openclaw} & stream-wrapper session ID equals the session file & stream contexts (count) \\
Pi\tablesource{piagent} & SDK session ID & completed assistants (count) \\
DeepSeek\tablesource{deepseekharness} & runner session ID & completed assistants (count) \\
Stirrup\tablesource{stirrup} & session ID from the native journal & \code{stop} / \code{tool\_calls} responses (count) \\
\bottomrule
\end{tabularx}
\end{table}

\subsection{Audits and training eligibility}
\label{app:audits}

Some failures leave a trajectory that looks complete. \roller{} audits the available evidence before pruning
auxiliary traces and records the results under \code{runtime} in \file{info.json}.

The \emph{thinking audit} flags visible reasoning after \code{off} was requested and reports
\code{not\_observable} when \flag{no-keep-thinking} suppresses that evidence. The \emph{wire-effort audit}
compares the requested level with controls in recorded requests, including retries. For Pi, DeepSeek, Hermes
and OpenCode, native request snapshots can supply evidence when the wire trace is absent. The result is
\code{unverified} when requests are present but their controls cannot establish the exact level.

The \emph{completion audit} uses the final ACTF step, available native events, and model response metadata to
flag a token-limit stop or an empty final answer without tool calls. Its outcomes are \code{incomplete},
\code{not\_flagged} and \code{not\_observable}. The \emph{model-identity audit} compares the model name in the
request with the names declared by the responses and records \code{name\_mismatch} when they differ.

\Cref{tab:eligibility} lists the reasons used by the SFT export policy. Audit exclusions preserve the rollout's
execution status and any benchmark score. A model name mismatch produces a warning because providers can report
aliases. The run summary counts linear, segmented and excluded samples. Eligible compacted samples go to
\file{sft\_segments.jsonl} with their supervision boundaries. XTuner export converts those boundaries to
per-message loss masks and combines the segments and linear records in one dataset.

\begin{table}[tbp]
\caption{Sample exclusions under the SFT export policy. An eligible sample yields a linear record or
context-window segments according to its captured history (\cref{sec:segments}).}
\label{tab:eligibility}
\vskip 0.06in
\centering
\scriptsize
\setlength{\tabcolsep}{3pt}
\renewcommand{\arraystretch}{1.1}
\begin{tabularx}{\linewidth}{@{}>{\raggedright\arraybackslash}p{0.33\linewidth}
  >{\raggedright\arraybackslash}X@{}}
\toprule
\tablehead
\textbf{Code} & \textbf{Trigger} \\
\midrule
\code{execution\_failed} & the rollout failed or timed out \\
\code{conversion\allowbreak\_unavailable} & ACTF or SFT conversion was unavailable, for example because
a context reset lacked captured Calls or the requests used programmatic tool definitions \\
\code{thinking\_effort\allowbreak\_mismatch} & visible reasoning after the level \code{off} was requested \\
\code{wire\_effort\allowbreak\_mismatch} & an outgoing request carried a different reasoning level than requested \\
\code{token\_limit} & the final response stopped on a token limit \\
\code{empty\_final\allowbreak\_answer} & the final response has neither text nor tool calls \\
\bottomrule
\end{tabularx}
\end{table}

\FloatBarrier
\section{Supplementary Experimental Results}
\label{app:experiments}

\subsection{Reasoning-level audit}
\label{app:thinking}

The results come from runs conducted in September and October 2026 with the Roller and Harness versions in use
at collection time. The reasoning-level matrix used Roller commit \code{af681e7} (PR \#64). The smoke test and
main matrix used \flag{wire-trace actf}. The control used \flag{wire-trace full}. All three phases retained visible
reasoning and used zero Roller retries.

A reasoning level passes through the CLI, the Harness configuration, the Harness's own request builder, any
gateway, and the provider; any of them can drop or rewrite it. We ran each of 7 Harnesses with each of 7 models
(\code{qwen3.8-flash}, \code{glm-5.3-flash}, \code{kimi-k2.7-code-highspeed}, \code{deepseek-v4-flash},
\code{gpt-5.6-luna}, \code{claude-opus-4-8}, \code{gemini-3.8-flash}) at four levels
(\code{off}, \code{low}, \code{high}, \code{max}). Each cell used the same deterministic task definition. The
agent had to write a marker to \file{outputs/answer.txt}, read it back and reply with the marker. A full pass
requires task success, a verified reasoning level and an eligible training record. \Cref{tab:thinking} gives
the verdict counts for all three phases.

\begin{table}[tbp]
\caption{Historical reasoning audit at Roller commit \code{af681e7}. The main matrix spans seven Harnesses,
seven models and four reasoning levels, with one deterministic tool task per cell and zero retries.
PASS requires task success, verified request settings and an eligible training record. The last column gives
SFT rows written divided by eligible samples.}
\label{tab:thinking}
\vskip 0.06in
\centering
\scriptsize
\setlength{\tabcolsep}{2.6pt}
\begin{tabular}{@{}l c c cccccccc c@{}}
\toprule
& & & \multicolumn{8}{c}{\textbf{Audit verdicts}} & \\
\cmidrule(lr){4-11}
\tablehead
\textbf{Stage} & \textbf{Capture} & \textbf{Cells} &
\textbf{PASS} & \textbf{S} & \textbf{U} & \textbf{T} & \textbf{R} &
\textbf{I} & \textbf{H} & \textbf{X} & \textbf{SFT rows / eligible} \\
\midrule
Smoke        & \code{actf} & 7   & 5  & 0 & 2  & 0 & 0 & 0  & 0  & 0  & 7 / 7 \\
Main matrix  & \code{actf} & 196 & 82 & 8 & 23 & 1 & 3 & 21 & 28 & 30 & 106 / 106 \\
Control (Qwen) & \code{full} & 28 & 20 & 1 & 2  & 0 & 0 & 0  & 4  & 1  & 22 / 22 \\
\bottomrule
\end{tabular}
\par\smallskip
\begin{minipage}{0.91\linewidth}
\raggedright\scriptsize
S = explicit reasoning conflict, U = unverified reasoning level, T = task mismatch,
R = unsupported SFT representation, I = timeout, H = preflight rejection, X = execution error.
The U verdict can accompany an otherwise eligible training record.
\end{minipage}
\end{table}

Of the 196 main cells, 82 passed the full check and all 8 explicit conflicts were excluded from the SFT
aggregate. Seven conflicts showed visible reasoning after \code{off} was requested, across OpenCode, DeepSeek,
Pi and OpenClaw. In the eighth, OpenClaw 2026.7.1 sent all three DeepSeek V4 Flash requests with \code{high}
when the run requested \code{low} (\cref{fig:cases}, panel (a)). The exact level was unverified in 23 cells
(14 Hermes and 9 OpenClaw).
Two OpenClaw cells in the full-wire control remained unverified because the request showed only that
reasoning was enabled.

The four task or conversion issues in \cref{fig:thinking} comprise one task mismatch and three Codex cells
whose \code{additional\_tools} definitions prevented SFT conversion (\cref{fig:cases}, panel (c)). Another
51 cells timed out or failed during execution (21 and 30, respectively). Preflight rejected the 28 cells
that requested \code{max} on Hermes or OpenCode, or \code{off} on Codex or Claude Code.
The main matrix produced 106 SFT rows, one per eligible sample. A sample with an unverified reasoning level
can still be eligible for export. The smoke and full-wire control also wrote one row per eligible sample
(\cref{tab:thinking}). A scan of all artifacts found no plaintext credential. A further 39 deterministic
regression checks cover the same boundaries without a model.

\subsection{Collection statistics}
\label{app:scale}

\Cref{tab:scale} describes the two teacher runs used for training in \cref{sec:exp-sft}. GLM-5.3-Flash on
OpenClaw started 493 of 1,000 animation tasks and 225 of 400 HTML slide tasks. When the runs were stopped
manually, \roller{} cancelled 682 queued rollouts and finalized all 718 that had started. Of the resulting
samples, 700 had an ACTF trace and 642 had a saved deliverable. The median animation trace had 13 steps
and 408.6k tokens. The median slide trace had 15 steps and 245.1k tokens, and the longest reached 147 steps.
Of the 135 failed rollouts, 78 hit API rate limits. During the overlap, the host ran up to 72 containers at once.

\begin{table}[!htbp]
\caption{Teacher collections on 23 October 2026 with GLM-5.3-Flash \citep{glm53flash} and OpenClaw
2026.7.1 \citep{openclaw}. Both runs were stopped manually. Failure categories come from the final
log entry. Token totals include prompts and completions across ACTF steps.}
\label{tab:scale}
\vskip 0.06in
\centering
\scriptsize
\setlength{\tabcolsep}{4pt}
\begin{tabular}{@{}l rr@{}}
\toprule
\tablehead
\textbf{Collection statistic} & \textbf{Animation} & \textbf{Slides} \\
\midrule
Tasks planned & 1,000 & 400 \\
Rollouts started & 493 & 225 \\
Queued rollouts cancelled & 507 & 175 \\
Samples with ACTF & 476 & 224 \\
Samples with a deliverable & 443 & 199 \\
\addlinespace[1.5pt]
Failed rollouts & 91 & 44 \\
\quad API rate limit & 45 & 33 \\
\quad network error / quota exhausted & 15 / 0 & 0 / 2 \\
\quad model timeout & 2 & 1 \\
\quad empty log / unclassified & 25 / 4 & 5 / 3 \\
\addlinespace[1.5pt]
ACTF steps per sample (median / mean / max) & 13 / 18.8 / 105 & 15 / 30.3 / 147 \\
Tokens per sample, median & 408.6k & 245.1k \\
Output tokens per sample, median & 63.4k & 35.9k \\
Tokens, total & 213.9M & 62.3M \\
Wall time per rollout, median & 1,561 s & 816 s \\
Duration of the run & 4.99 h & 3.07 h \\
\bottomrule
\end{tabular}
\par\smallskip
\begin{minipage}{0.91\linewidth}
\raggedright\scriptsize
With both runs overlapping, the host ran up to 72 containers (mean 63.2). Container memory peaked
at 135.2 GB combined and 2.3 GB for the largest container. The Roller process used up to 10.8 GB.
\end{minipage}
\end{table}

\subsection{Teacher reasoning and deliverables}
\label{app:reasoning}

\begin{figure}[!htbp]
\centering
\includegraphics[width=0.75\columnwidth]{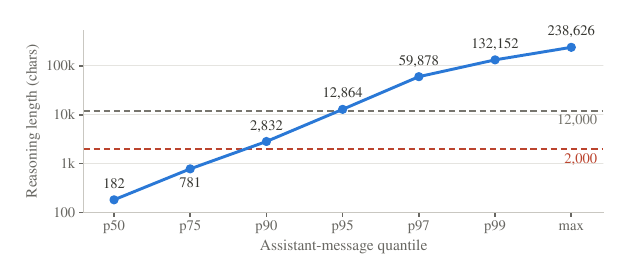}
\caption{Reasoning length over 23,235 assistant messages in the teacher trajectories. The vertical axis is
logarithmic. Dashed lines mark the 2,000- and 12,000-character blanking thresholds in \cref{tab:sft}.}
\label{fig:reasoning}
\end{figure}

In an earlier round trained on the records as collected, the model spent its token budget on reasoning in
14 of 20 held-out tasks and returned neither content nor tool calls. \Cref{fig:reasoning} shows a long tail in
the teacher's \code{reasoning\_content}. Across 23,235 assistant
messages, the median length is 182 characters, the 97th percentile is 59,878 and the maximum is 238,626.
The 2,000-character cutoff falls between the 75th and 90th percentiles, and the 12,000-character cutoff
falls between the 90th and 95th. Messages above 4,000 characters are 8\% of all messages and hold 91.7\%
of the reasoning text.
Degenerate reasoning sits in only 3.4\% of messages but in 75.5\% of samples, so dropping whole samples would have
discarded most of the data.

The 2,000-character animation arm scored 9.50 and produced a median of 192 HTML tags.
The base model scored 9.89 with a median of 140 tags, and
the teacher's median was 193 on the training tasks.

\FloatBarrier
\section{Diagnostic Case Studies}
\label{sec:cases}

\begin{figure}[H]
\centering
\begin{tcbraster}[raster columns=2, raster equal height=rows, raster column skip=7pt, raster row skip=7pt]
\begin{casebox}{(a) Reasoning level rewritten}{rorange}
\turn{Run} DeepSeek V4 Flash on OpenClaw 2026.7.1. The marker task passed.\par\smallskip
\turn{Evidence} The run requested \tool{low}. All three outgoing requests carried \tool{high}.\par\smallskip
\turn{Outcome} The wire-effort audit excluded the sample from the SFT aggregate.
\end{casebox}
\begin{casebox}{(b) Search failed throughout the benchmark}{rblue}
\turn{Run} Two 60-task WildClawBench runs on OpenClaw. Graders still produced scores.\par\smallskip
\turn{Evidence} All 131 search calls returned \tool{fetch failed}.\par\smallskip
\turn{Outcome} The tool observations exposed missing upstream search setup in the adapter.
\end{casebox}
\begin{casebox}{(c) Codex tools blocked SFT export}{rviolet}
\turn{Run} Three Codex cells with GPT-5.6-Luna passed the task and request checks.\par\smallskip
\turn{Evidence} The captured requests contain \tool{additional\_tools}, preserved in Codex's ACTF tool capture.\par\smallskip
\turn{Outcome} SFT conversion was unavailable for these programmatic tool definitions.
\end{casebox}
\begin{casebox}{(d) Final answer truncated}{rorange}
\turn{Run} SWE-Bench Pro, task index 70, with DeepSeek V4 Pro on OpenClaw.\par\smallskip
\turn{Evidence} Both attempts ended on \tool{length}. The empty patch scored zero.\par\smallskip
\turn{Outcome} The \tool{length} stop meets Roller's \tool{token\_limit} exclusion rule for SFT.
\end{casebox}
\end{tcbraster}
\caption{Four diagnostic cases from recorded rollouts. Outgoing requests, tool observations, captured tool
definitions and stop reasons explain the benchmark or training outcome in each case.}
\label{fig:cases}
\end{figure}

The cases in \cref{fig:cases} show what each kind of evidence adds to the benchmark result. Request
capture exposes a changed reasoning setting and tool definitions that the training format cannot
represent. Tool observations reveal a missing adapter setting despite completed grading
(\cref{sec:bench}). The stop reason explains why the SWE-Bench Pro rollout produced an empty patch
and how Roller's completion audit would classify that outcome.

\end{document}